\documentclass[runningheads]{llncs}

\usepackage{eccv}
\usepackage{eccvabbrv}

\usepackage[export]{adjustbox}
\usepackage{amsfonts}
\usepackage{amsmath}
\usepackage{amssymb}
\usepackage{microtype}
\usepackage{multirow}
\usepackage{nicefrac}
\usepackage{pifont}
\usepackage{tabularx}
\usepackage{url}
\usepackage{wrapfig}
\usepackage{xcolor}
\usepackage{xspace}

\usepackage{graphicx}
\usepackage{booktabs}

\usepackage[accsupp]{axessibility}  %

\usepackage{hyperref}

\usepackage{orcidlink}

\newcommand{\cmark}{\ding{51}}%
\newcommand{\xmark}{\ding{55}}%

\DeclareMathOperator*{\argmax}{arg\,max}
\DeclareMathOperator*{\argmin}{arg\,min}

\newcommand{\NAME}{OVDiff\xspace}

\begin{document}

\title{Diffusion Models for Open-Vocabulary Segmentation}

\author{Laurynas Karazija \and
Iro Laina \and
Andrea Vedaldi \and
Christian Rupprecht}

\authorrunning{L.~Karazija et al.}

\institute{Visual Geometry Group, Department of Engineering Science,
University of Oxford
\email{\{laurynas,iro,vedaldi,chrisr\}@robots.ox.ac.uk}}

\maketitle

\begin{abstract}
Open-vocabulary segmentation is the task of segmenting anything that can be named in an image. 
Recently, large-scale vision-language modelling has led to significant advances in open-vocabulary segmentation, but at the cost of gargantuan and increasing training and annotation efforts.
Hence, we ask if it is possible to use \emph{existing} foundation models to synthesise on-demand efficient segmentation algorithms for specific class sets, making them applicable in an open-vocabulary setting without the need to collect further data, annotations or perform training.
To that end, we present \NAME{}, a novel method that leverages generative text-to-image diffusion models for unsupervised open-vocabulary segmentation. 
\NAME{} synthesises support image sets for arbitrary textual categories, creating for each a set of prototypes representative of both the category and its surrounding context (background).
It relies solely on pre-trained components and outputs the synthesised segmenter directly, without training.
Our approach shows strong performance on a range of benchmarks, obtaining a lead of more than 5\% over prior work on PASCAL VOC.
\end{abstract}

\begin{figure}[t]
    \centering
    \includegraphics[width=\textwidth]{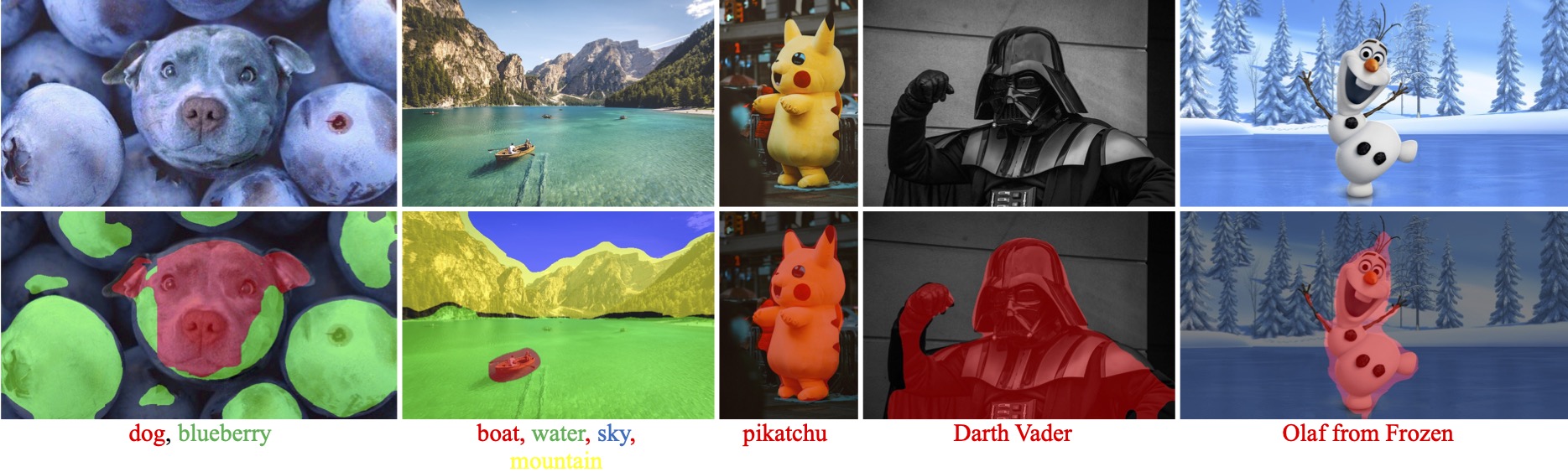}
    \captionof{figure} {\NAME{} is an open-vocabulary segmentation method that, given an image and a free-form set of class names, can segment any user-defined classes. It is fully automatic and does not require any further training. \vspace{2em}}
    \label{fig:teaser}
\end{figure}
\section{Introduction}\label{sec:intro}

Open-vocabulary semantic segmentation is the task of segmenting images into regions matching several free-form textual categories.
As the field of Computer Vision moves towards large-scale general-purpose models, open-vocabulary ``foundation'' models have similarly emerged.
Yet, the development of ones suitable for dense localisation tasks such as semantic segmentation incurs both enormous training costs and requires expensive mask annotations.
Instead, we show that the open-vocabulary segmentation task can be effectively tackled starting from a set of frozen foundation models, without requiring additional data or even fine-tuning.

In order to do so, we introduce \NAME{}, a method that turns existing foundation models into a ``factory'' of image segmenters, \ie, using foundation models to synthesise on-demand a segmenter for any new concepts specified in natural language.
Thus, \NAME{} can be used for open-vocabulary segmentation, where it achieves state-of-the-art results in standard benchmarks.
Moreover, once synthesised, the segmenters can be efficiently applied to any number of images and easily extended to new categories.

Specifically, segmenting an image using \NAME{} can be done in three steps: \textit{generation}, \textit{representation}, and \textit{matching}.
Given a textual prompt, \NAME{} uses an off-the-shelf text-to-image generator like StableDiffusion~\cite{rombach2022high} to \textit{generate} a support set of images.
In the representation step, we use a feature extractor (that can be the same network as in the generation step) to extract feature prototypes that represent the textual category.
Finally, we use simple nearest-neighbour \textit{matching} scheme to segment the target image using the prototypes computed in the previous step.

This approach differs from prior work that largely approaches the problem in either of two ways. 
Starting from multi-modal representations (\eg, CLIP~\cite{radford2021learning}) to bridge vision and language, the first way relies on labelled data 
to fine-tune image-level representations for the segmentation task.
Hence, in line with the zero-shot setting~\cite{bucher2019zero}, these methods require costly dense annotations for some known categories while also extending the segmentation to unseen categories by incorporating language.

The second category of prior work~\cite{xu2022groupvit,ren2023viewco,xu2023learning,lou2022segclip,mukhoti2022open,cha2022learning} observes that large-scale vision-language models such as CLIP have a limited understanding of the positioning of objects within an image and extend these models with additional grouping mechanisms for better localisation using only image-level captions, but no mask supervision.
This, however, requires expensive additional contrastive training at scale.
Additionally, most methods resort to heuristics to segment the background (\ie, leave some pixels unlabelled), as it often cannot be described as a textual category. 
The usual approach is to threshold the similarities to all categories.
Finding an appropriate threshold, however, can be challenging and may vary depending on the image, often resulting in imprecise object boundaries.
Effectively handling the background remains an open issue.

Our three-step approach departs substantially from both of these schemes. We show that large-scale text-to-image generative models, such as StableDiffusion~\cite{rombach2022high}, can help bridge the vision-and-language gap without the need for annotations or costly training.
Furthermore, diffusion models also produce latent spaces that are semantically meaningful and well-localised.
This solves a second problem:  multi-modal embeddings are difficult to learn and often suffer from ambiguities and differences in detail between modalities.
Instead, our approach can use unimodal features for open-vocabulary segmentation, which offers several advantages.
Firstly, as text-to-image generators encode a distribution of possible images, this offers a means to deal with intra-class variation and captures the ambiguity in textual descriptions.
Secondly, the generative image models encode not only the visual appearance of objects but also provide contextual priors, which we use for direct background segmentation.

This work presents a simple framework that achieves state-of-the-art performance across open-vocabulary segmentation benchmarks. It combines several off-the-shelf pre-trained networks into a segmenter ``factory'' that segments images into arbitrary textual categories in three simple steps. \NAME{} requires no additional data, mask supervision, nor fine-tuning. 
To summarise, we make the following core contributions:
(1) We introduce a method to use pre-trained diffusion models for the task of open-vocabulary segmentation, that requires no additional data, mask supervision, or fine-tuning.
(2) We propose a principled way to handle backgrounds by forming prototypes from contextual priors built into text-to-image generative models.
(3) A set of additional techniques for further improving performance, such as multiple prototypes, category filtering and "stuff" filtering.

\section{Related work}\label{sec:rel_works}

\paragraph{Zero-shot open-vocabulary segmentation.}
Open-vocabulary semantic segmentation is a relatively new problem and is typically approached in two ways. 
The first line of work poses the problem as ``zero-shot'', \ie, segmenting unseen classes after training on a set of observed classes with dense annotations.
Early approaches~\cite{bucher2019zero,li2020consistent,gu2020context,cheng2021sign}  %
explore generative networks to sample features using conditional language embeddings for classes. 
In \cite{xian2019semantic,li2021language} image encoders are trained to output dense features that can be correlated with word2vec~\cite{mikolov2013distributed} and CLIP~\cite{radford2021learning} text embeddings. %
Follow-up works~\cite{ghiasi2022scaling,liang2022open,ding2022decoupling,xu2022simple} approach the problem in two steps, predicting class-agnostic masks and aligning the embeddings of masks with language.
IFSeg~\cite{yun2023ifseg} generates synthetic feature maps by pasting CLIP text embeddings into a known spatial configuration to use as additional supervision. 
Different from our approach, all these works rely on mask supervision %
for a set of known classes. %

The second line of work eliminates the need for mask annotations and instead aims to align image regions with language using only image-text pairs. 
This is largely enabled by recent advancements in large-scale vision-language models~\cite{radford2021learning}.
Some methods introduce internal grouping mechanisms such as hierarchical grouping~\cite{xu2022groupvit,ren2023viewco,wysoczanska2024clip}, slot-attention~\cite{xu2023learning}, or cross-attention to learn cluster centroids~\cite{liu2022open,lou2022segclip}.
Assignment to language queries is performed at group level.
Another line of work~\cite{zhou2022maskclip,mukhoti2022open,cha2022learning,ranasinghe2022perceptual} aims to learn dense features that are better localised when correlated with language embeddings at pixel level. 
With the exception of \cite{ranasinghe2022perceptual,zhou2022maskclip,wysoczanska2024clip}, thresholding is often required to determine the background during inference. Alternatively, a curated list of background prompts can be used~\cite{ranasinghe2022perceptual}.

Our method falls into the second category. 
However, in contrast to prior work, we leverage a generative model to translate language queries to pre-trained image feature extractors without further training. 
We also segment the background directly, without relying on thresholding or curated list of background prompts.
A closely related approach to ours is ReCO~\cite{shin2022reco}, where CLIP is used for image retrieval compiling a set of exemplar images from ImageNet for a given language query, which is then used for co-segmentation.
In our method, the shortcoming of an image database is addressed by synthesising data on-demand. %
Furthermore, instead of co-segmentation, we leverage the cross-attention of the generator to extract objects.
Instead of similarity of support images, we use diverse samples and both foreground and contextual backgrounds.
Follow up works~\cite{barsellotti2024fossil,barsellotti2024training} to \NAME{} exchange contextual prior for backgrounds with compiling a database of prototypes.

\paragraph{Diffusion models.}
Diffusion models~\cite{sohl2015deep,ho2020denoising,song2021score} are a class of generative methods that have seen tremendous success in text-to-image systems such as DALL-E~\cite{ramesh2022hierarchical}, Imagen~\cite{saharia2022photorealistic}, and Stable Diffusion~\cite{rombach2022high}, trained on Internet-scale data such as LAION-5B~\cite{schuhmann2022laion}.
The step-wise generative process and the language conditioning make pre-trained diffusion models attractive also for discriminative tasks.
They have been recently used in few-shot classification~\cite{zhang2023prompt}, few-shot segmentation~\cite{baranchuk2022label} and panoptic segmentation~\cite{xu2023open}, and to generate pairs of images and segmentation masks~\cite{li2023grounded}.
However, these methods rely on dense manual annotations to associate diffusion features with the desired output.

Annotation-free discriminative approaches such as~\cite{li2023your,clark2023text,udandarao2023sus} use pre-trained diffusion models as zero-shot classifiers.
DiffuMask~\cite{wu2023diffumask} uses prompt engineering to synthesise a dataset of ``known'' and ``unseen'' categories and trains a closed-set segmenter with masks obtained from the cross-attention maps of the diffusion model.
DiffusionSeg~\cite{ma2023diffusionseg} uses DDIM inversion~\cite{song2021score} to obtain feature maps and attention masks of object-centric images to perform unsupervised object discovery, but relies on ImageNet labels and is not open-vocabulary.
Our approach also leverages the rich semantic information present in diffusion models for segmentation; unlike these methods, however, it is open-set and does not require further training.

\paragraph{Unsupervised segmentation.}
Our work is also related to unsupervised segmentation approaches.  
While early works relied on hand-crafted priors~\cite{cheng2015global,wei2012geodesic,zhang2018deep,zeng2019multi,nguyen2019DeepUSPS} later approaches leverage feature extractors such as DINO~\cite{caron2021emerging} and perform further analysis of these methods \cite{wang2022selfsupervised,melas-kyriazi2022deep,simeoni2021localizing,simeoni2022unsupervised,hamilton2022unsupervised,shin2022selfmask,wang2023cutler,wang2022freesolo}. 
Some approaches make use of generative methods, usually GANs, %
to separate images in foreground and background layers~\cite{bielski2019emergence,chen2019unsupervised,benny2020onegan,bielski2022move} or analyse latent structure to induce known foreground-background changes~\cite{voynov2021object,melas-kyriazi2022finding} to synthesise a training dataset with labels. Some works explore interaction with different modalities such as optical flow~\cite{choudhury+karazija22gwm,karazija22unsupervised} or depth~\cite{bowen2022dimensions}.
Largely focused on unsupervised saliency prediction,  %
these methods are class-agnostic and do not incorporate language.

\newcommand{\sset}{\mathcal{S}}  %
\newcommand{\ssi}{S_n}       %
\newcommand{\ssc}{|\sset|}   %

\section{Method}\label{sec:method}

We present \NAME{}, a method for open-vocabulary segmentation, \ie, semantic segmentation of any category described in natural language.
We achieve this goal in three steps:
(1) we leverage text-to-image generative models to \emph{generate} a set of images representative of the described category,
(2) use these to ground \emph{representations} from off-the-shelf pretrained feature extractors, and
(3) \emph{match} these against input image features to perform segmentation.

\subsection{\NAME{}: Diffusion-based open-vocabulary segmentation}

Our goal is to devise an algorithm which, given a new vocabulary of categories $c_i \in \mathcal{C}$ formulated as natural language queries, can segment any image against it.
Let $I \in \mathbb{R}^{H \times W \times 3}$ be an image to be segmented. 
Let
$
\Phi_v: \mathbb{R}^{H \times W \times 3} \to \mathbb{R}^{H'W' \times D}
$
be an off-the-shelf visual feature extractor and
$
\Phi_t: \mathbb{R}^{d_t} \to \mathbb{R}^D
$
a text encoder.
Assuming that image and text encoders are aligned, one can achieve segmentation by simply computing a similarity function, for example, the cosine similarity $s(\Phi_v(I), \Phi_t(c_i))$, with $s(x, y) = \frac{x^T y}{\|x\| \|y\|}$, between the encoded image $\Phi_v(I)$ and an encoding of a class label $c_i$.
To meaningfully compare different modalities, image and text features must lie in a shared representation space, which is typically learned by jointly training $\Phi_v$ and $\Phi_t$ using image-text or image-label pairs~\cite{radford2021learning}.

We propose two modifications to this approach.
First, we observe that it is better to compare representations of the \textit{same} modality than across vision and language modalities.
We thus replace $\Phi_t(c_i)$ with a $D$-dimensional \textit{visual} representation $\bar{P}$ of class $c_i$, which we refer to as a \textit{prototype}.
In this case, the same feature extractor can be used for both prototypes and target images; thus, their comparison becomes straightforward and does not necessitate further training. Second, we propose utilising \emph{multiple} prototypes per category instead of a single class embedding.
This enables us to accommodate intra-class variations in appearance, and, as we explain later, it also allows us to exploit contextual priors, which in turn help to segment the background.

Our approach, thus, proceeds in three steps: (1) a set of support images is sampled based on vocabulary $\mathcal{C}$, (2) a set of prototypes $\mathcal{P}$ is calculated, and (3) a set of images $\{I_1,I_2\dots\}$ is segmented against these prototypes. 
We observe that in practical applications, whole image collections are processed using the same vocabulary, as altering the set of target classes for individual images in an informed way would already require some knowledge of their contents. 
Steps (1) and (2) are, thus, performed very infrequently, and their cost is heavily amortised.
Next, we detail each step.

\begin{figure*}[t]
    \centering
    \includegraphics[width=0.93\textwidth]{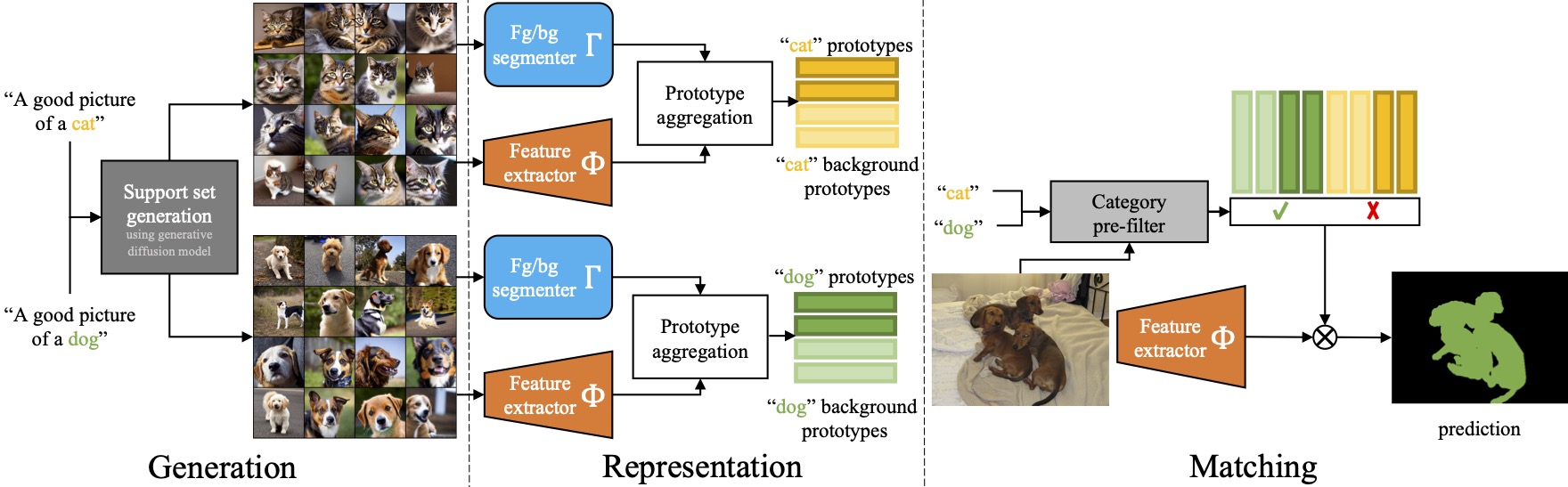}
    \caption{OVDiff overview. Prototype sampling: text queries are used to sample a set of support images which are further processed by a feature extractor and a segmenter forming positive and negative (background) prototypes. Segmentation: image features are compared against prototypes.%
    The CLIP filter removes irrelevant prototypes based on global image contents.}
    \label{fig:method}
\end{figure*}
\subsection{Support set generation}

To construct a set of prototypes, the first step of our approach is to sample a support set of images representative of each category $c_i$.
This can be accomplished by leveraging pretrained text-conditional generative models. 
Sampling images from a generative model, as opposed to a curated dataset of real images, aligns well with the goals of open-vocabulary segmentation as it enables the construction of prototypes for \textit{any} user-specified category or description, even those for which a manually labelled set may not be readily available (\eg, $c_i=$ ``\texttt{donut with chocolate glaze}'').

Specifically, for each query $c_i$, we define a prompt ``\texttt{A good picture of a $\langle c_i \rangle$}'' and generate a small batch of $N$ support images $\sset = \{ S_1, S_2, \dots, S_N \mid S_n \in \mathbb{R}^{hw \times 3}\}$ of height $h$ and width $w$ using Stable Diffusion~\cite{rombach2022high}.

\subsection{Representing categories}

Na\"{\i}vely, prototypes $\bar{P}_{c_i}$ could be constructed by averaging all features across all images for class $c_i$.
This is unlikely to result in good prototypes because not all pixels in the sampled images correspond to the class specified by $c_i$.
Instead, we propose to extract the class prototypes as follows.
\paragraph{Class prototypes.}
Our approach generates two sets of prototypes, positive and negative, for each class.
Positive prototypes are extracted from image regions that are associated with $\langle c_i \rangle$, while negative prototypes represent ``background'' regions.
Thus, to obtain prototypes, the first step is segmenting the sampled images into foreground and background.
To identify regions most associated with $c_i$, we use the fact that the layout of a generated image is largely dependent on the cross-attention maps of the diffusion model~\cite{hertz2022prompt}, \ie, pixels attend more strongly to words that describe them.
For a given word or description (in our case $c_i$), one can generate a set of attribution maps $\mathcal{A} = \{A_1, A_2, \dots, A_N \mid A_n \in \mathbb{R}^{hw}\}$, corresponding to the support set $\mathcal{S}$, by summing the cross-attention maps across all layers, heads, and denoising steps of the network~\cite{tang2022daam}.

Yet, thresholding these attribution maps may not be optimal for segmenting foreground/background, as they are often coarse or incomplete, and sometimes only parts of objects receive high activation. To improve segmentation quality, we propose to optionally leverage an unsupervised instance segmentation method $\Gamma$.
Unsupervised segmenters are not vocabulary-aware and may produce multiple binary object proposals.
We denote these as $\mathcal{M}_n = \{ M_{nr} \mid M_{nr} \in \{0,1\}^{hw} \}$, where $n$ indexes the support images and $r$ indexes the object masks (including a mask for the background). We thus construct a promptable extension of $\Gamma$ segmenter to select appropriate proposals for foreground and background: 
for each image, we select from $\mathcal{M}_n$ the mask with the highest (lowest) average attribution as the foreground (background):
\begin{equation}
    M^{\operatorname{fg}}_n = \argmax_{M \in \mathcal{M}_n} \frac{M^\top A_n}{M^\top M}, \quad
    M^{\operatorname{bg}}_n = \argmin_{M \in \mathcal{M}_n} \frac{M^\top A_n}{M^\top M}.
\end{equation}

\paragraph{Prototype aggregation.} We can compute prototypes $P_n^{\operatorname{g}}$ for foreground and background regions ($\operatorname{g} \in \{\operatorname{fg}, \operatorname{bg} \}$) as
\begin{equation}\label{eq;prototype}
    P_n^{\operatorname{g}} = \frac{(\hat{M}^{\operatorname{g}}_n)^\top  \Phi_v(S_n)}{m^{\operatorname{g}}_n} \, \in \mathbb{R}^D,
\end{equation}
where $\hat{M}^{\operatorname{g}}_n$ denotes a resized version of $M^{\operatorname{g}}_n$ that matches the spatial dimensions of $\Phi_v(S_n)$, and $m^{\operatorname{g}}_n = (\hat{M}^{\operatorname{g}}_n)^\top \hat{M}^{\operatorname{g}}_n$ counts the number of pixels within each mask.
In other words, prototypes are obtained by means of an off-the-shelf pretrained feature extractor and computed as the average feature within each mask.

We refer to these as \textit{instance} prototypes because they are computed from each image individually, and each image in the support set can be viewed as an instance of class $c_i$.

In addition to instance prototypes, we found it helpful to also compute \textit{class-level} prototypes $\bar{P}^{\operatorname{g}}$  by averaging the instance prototypes weighted by their mask sizes as
$
\bar{P}^{\operatorname{g}}
=
\sum_{n=1}^N{m^{\operatorname{g}}_n P_n^{\operatorname{g}}}
/
\sum_{n=1}^N{m^{\operatorname{g}}_n}
$.

Finally, we propose to augment the set of class and instance prototypes %
using $K$-Means clustering of the masked features to obtain \textit{part-level} prototypes. %
We perform spatial clustering separately on foreground and background regions and take each cluster centroid as a prototype $P^{\operatorname{g}}_k$ with $1 \le k \le K$.
The intuition behind this is to enable segmentation at the level of parts, support greater intra-class variability, and a wider range of feature extractors that might not be scale invariant. %

We consider the union of all these feature prototypes:
\begin{equation}
\mathcal{P}^{\operatorname{g}} =
\bar{P}^{\operatorname{g}} \cup
\{ P^{\operatorname{g}}_n \mid 1 \le n \le N\} \cup
\{ P^{\operatorname{g}}_k \mid 1 \le k \le K\}
\end{equation}
for $\,\operatorname{g} \in \{\operatorname{fg},\operatorname{bg} \}$, and associate them with a single category.

We note that this process is repeated for each $c_i \in \mathcal{C}$ and we hereby refer to $\mathcal{P}^{\operatorname{fg}}$ (and $\mathcal{P}^{\operatorname{bg}}$) as $\mathcal{P}^{\operatorname{fg}}_{c_i}$ ($\mathcal{P}^{\operatorname{bg}}_{c_i}$), \ie, as the foreground (background) prototypes of class $c_i$.

Since $\mathcal{P}^{\operatorname{fg}}_{c_i}$ ($\mathcal{P}^{\operatorname{bg}}_{c_i}$) depend only on class $c_i$, they can be precomputed, and the set of classes can be dynamically expanded without the need to adapt existing prototypes.

\subsection{Segmentation via prototype matching}

To perform segmentation of any target image $I$ given a vocabulary $\mathcal{C}$, we first extract image features using the same visual encoder $\Phi_v$ used for the prototypes.
The vocabulary is expanded with an additional background class $\hat{\mathcal{C}} = \{c_{\operatorname{bg}}\}\cup \mathcal{C}$, for which the positive (\textit{foreground}) prototype is the union of all \textit{background} prototypes in the vocabulary: $\mathcal{P}^{\operatorname{fg}}_{c_{\operatorname{bg}}}$ $ = \bigcup_{c_i \in \mathcal{C}} \mathcal{P}^{\operatorname{bg}}_{c_i}$.
Then, a segmentation map can simply be obtained by matching dense image features to prototypes using cosine similarity. 
A class with the highest similarity in its prototype set is chosen:
\begin{equation}\label{eq:segmentation}
M = \argmax_{c \in \hat{\mathcal{C}}}\max_{P \in \mathcal{P}^{\operatorname{fg}}_{c}}s(\Phi_v(I), P).
\end{equation}

\paragraph{Category pre-filtering.}
To limit the impact of spurious correlations that might exist in the feature space of the visual encoder, we introduce a pre-filtering process for the target vocabulary given image $I$.
Specifically, we leverage CLIP~\cite{radford2021learning} as a strong open-vocabulary classifier but propose to apply it in a multi-label fashion to constrain the segmentation to the subset of categories $\mathcal{C}' \subseteq \mathcal{C}$ that appear in the target image. 
First, we encode the target image and each category using CLIP.
Any categories that do not score higher than $\nicefrac{1}{|\mathcal{C}|}$ are removed from consideration, that is we keep the subset
$\{P^{\operatorname{g}}_{c'} \mid c' \in \mathcal{C}'\}$, $\operatorname{g} \in \{\operatorname{fg}, \operatorname{bg}\}$.
If more than $\eta$ categories are present, then the top-$\eta$ are selected.
We then form ``multi-label'' prompts as ``\texttt{$\langle c_a \rangle$ and $\langle c_b \rangle$ and ...}'' where the categories are selected among the top scoring ones taking into account all $2^{\eta}$ combinations.
The best-scoring multi-label prompt determines the final list of categories to be used in \Cref{eq:segmentation}.

\paragraph{``Stuff'' filtering.} 
Occasionally, $c_i$ might not describe a countable object category but an identifiable region in the image, \eg, \texttt{sky}, often referred to as a ``stuff'' class.
``Stuff'' classes warrant additional consideration as they might appear as background in images of other categories, \eg, \texttt{boat} images might often contain regions of \texttt{water} and \texttt{sky}.
As a result, the process outlined above might sample background prototypes for one class that coincide with the foreground prototypes of another.
To mitigate this issue, we introduce an additional filtering step to detect and reject such prototypes, when the full vocabulary, \ie, the set of classes under consideration, is known.
First, we only consider foreground prototypes for ``stuff'' classes.
Additionally, any negative prototypes of ``thing'' classes with high cosine similarity with any of the ``stuff'' class prototypes are simply removed.
In our experiments, we use ChatGPT~\cite{chatgpt} to automatically categorise a set of classes as ``thing'' or ``stuff''.

\begin{table}[t]
\centering \footnotesize
\caption{{Open-vocabulary segmentation}. Comparison of our approach, \NAME{}, to the state of the art (under the mIoU metric). Our results are an average of 5 seeds $\pm\sigma$. $^*$results from~\cite{cha2022learning}.
}
\label{tab:main_results}
\begin{tabular}{@{}l@{\hspace{-2pt}}c@{\hspace{2pt}}c@{\hspace{-4pt}}c@{\hspace{2pt}}c@{\hspace{1pt}}c@{}}
\toprule
\multirow{2}{*}{\textbf{Method}}    & \textbf{Support}  & \textbf{Further}                                               & \multirow{2}{*}{\textbf{VOC}}                     & \multirow{2}{*}{\textbf{Context}}                                  &  \multirow{2}{*}{\textbf{Object}}                  \\
& \textbf{Set} & \textbf{Training} & & & \\
\midrule
ReCo$^*$~\cite{shin2022reco}  & Real & \xmark                    & 25.1                    & 19.9                                     &  15.7                    \\

ViL-Seg~\cite{liu2022open}     & \xmark   & \cmark                    & 37.3                    & 18.9                                     &  -                       \\
MaskCLIP$^*$~\cite{zhou2022maskclip}  & \xmark   & \xmark              & 38.8                    & 23.6                                     &  20.6                    \\
TCL~\cite{cha2022learning}  & \xmark   & \cmark                        & 51.2                    & 24.3                                     &  30.4                    \\
CLIPpy~\cite{ranasinghe2022perceptual}   & \xmark   & \cmark           & 52.2                    & -                                        &  \underline{32.0}        \\
GroupViT~\cite{xu2022groupvit}    & \xmark   & \cmark                  & 52.3                    & 22.4                                     &  -                       \\
ViewCo~\cite{ren2023viewco}    & \xmark   & \cmark                     & 52.4                    & 23.0                                     &  23.5                    \\
SegCLIP~\cite{lou2022segclip}   & \xmark   & \cmark                    & 52.6                    & \underline{24.7}                         &  26.5                    \\
OVSegmentor~\cite{xu2023learning}   & \xmark   & \cmark                & 53.8        & 20.4                                     &  25.1                    \\
CLIP-DIY~\cite{wysoczanska2024clip} & \xmark & \xmark & \underline{59.9} & -- & 31.0 \\ 
\textbf{\NAME{}} {\footnotesize (-CutLER)}    & Synth.   & \xmark                                  & 62.8 & 28.6 & 34.9 \\
\textbf{\NAME{}}    & Synth.   & \xmark                                    & \textbf{66.3 $\pm$ 0.2} & \textbf{29.7 $\pm$ 0.3}                  &  \textbf{34.6 $\pm$ 0.3} \\
\midrule
\midrule
TCL~\cite{cha2022learning} {\footnotesize(+PAMR)} & \xmark   & \cmark & \underline{55.0}        & \underline{30.4}                         &  \underline{31.6}        \\
\textbf{\NAME{}} {\footnotesize(+PAMR)}  & Synth.   & \xmark             & \textbf{ 68.4 $\pm$ 0.2 } & \textbf{ 31.2 $\pm$ 0.4 }                  &  \textbf{ 36.2 $\pm$ 0.4 } \\ \bottomrule
\end{tabular}
\end{table}

\section{Experiments}\label{sec:experiments}
We evaluate \NAME{} on the open-vocabulary semantic segmentation task.
First, we consider different feature extractors and investigate how they can be grounded by leveraging our approach.
We then turn to comparisons of our method with prior work.
We ablate the components of \NAME{}, visualize the prototypes, and conclude with a qualitative comparison with prior works on in-the-wild images.

\paragraph{Datasets and implementation details.}

As the approach does not require further training of components, we only consider data for evaluation.
Following prior work~\cite{xu2022groupvit}, to assess the segmentation performance, we report mean Intersection-over-Union (mIoU) on validation splits of PASCAL VOC (VOC)~\cite{pascal-voc-2012}, PASCAL Context (Context)~\cite{mottaghi2014role} and COCO-Object (Object)~\cite{caesar2018cocostuff} datasets, with 20, 59, and 80 foreground classes, respectively.
These datasets include a background class to reflect a realistic setting of non-exhaustive vocabularies.
Context also contains both ``things'' and ``stuff'' classes.
We also evaluate without background on VOC, Context, ADE20K~\cite{zhou2017scene}, COCO-Stuff~\cite{caesar2018cocostuff} and Cityscapes~\cite{Cordts_2016_CVPR}, with 20, 59, 150, 171, and 19 classes, respectively,  but do not consider this a realistic setting as it relies on knowing which pixels cannot be described by a set of categories.
Similar to~\cite{cha2022learning,xu2022groupvit,xu2023learning}, we employ a sliding window approach.
We use two scales to aid with the limited resolution of off-the-shelf feature extractors with square window sizes of 448 and 336 and a stride of 224 pixels.
We set the size of the support set to $N=32$. For the diffusion model, we use Stable Diffusion v1.5; for unsupervised segmenter $\Gamma$, we employ CutLER~\cite{wang2023cutler}. 

\subsection{Grounding feature extractors}
Our method can be combined with \textit{any} pretrained visual feature extractor for constructing prototypes and extracting image features.
To verify this quantitatively, we experiment with various self-supervised ViT feature extractors (\cref{tab:features}): DINO~\cite{caron2021emerging}, MAE~\cite{he2022masked}, and CLIP~\cite{radford2021learning}.
We also use SD as a feature extractor.

We find that SD performs the best, though CLIP and DINO also show strong performance based on our experiments on VOC.
MAE shows the weakest performance, which may be attributed to its lack of semanticity~\cite{he2022masked}; yet it is still competitive with the majority of purposefully trained networks when employed as part of our approach.
We find that taking \textit{keys} of the second to last layer in CLIP yields better results than using patch tokens (CLIP token).
As feature extractors have different training objectives, we hypothesise that their feature spaces might be complementary. 
Thus, we also consider an ensemble approach.
In this case, the cosine distances formed between features of different extractors and respective prototypes are averaged.
The combination of SD, DINO, and CLIP performs the best.
We adopt this formulation for the main set of experiments.

\begin{figure*}[t]
    \centering
    \includegraphics[width=0.94\textwidth]{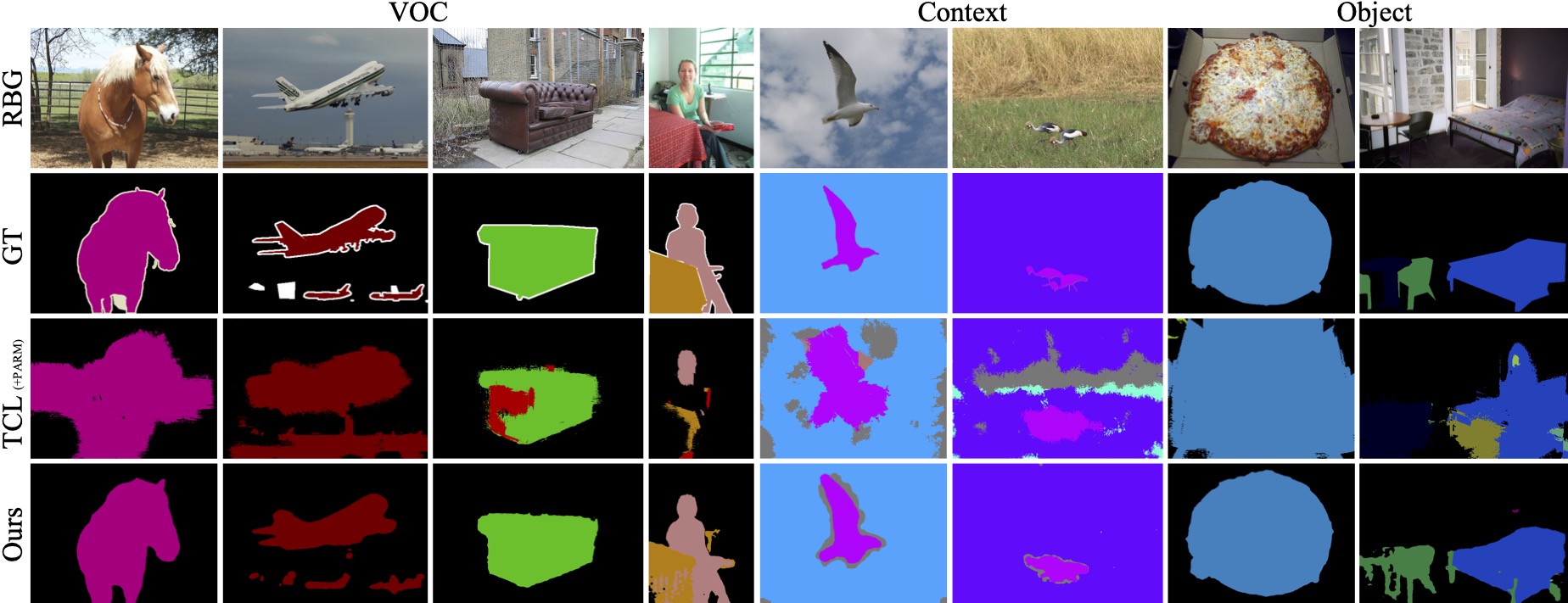}
    \caption{{Qualitative results}. \NAME{} in comparison to TCL {\footnotesize(+ PAMR)}. \NAME{} provides more accurate segmentations across a range objects and stuff classes with well defined object boundaries that separate from the background well.
    }
    \label{fig:main_qual}
\end{figure*}%
\subsection{Comparison to existing methods}
In \cref{tab:main_results}, we compare our method with prior work that does not rely on manual mask annotation on three datasets: VOC, Context, Object.
We include a brief overview of the methods in the supplement.
We find that our method compares favourably, outperforming other methods in all settings.
In particular, results on VOC show the largest margin, with more than 5\% improvement over prior work.

We also consider a version of our method, \NAME{} (-CutLER), that does not rely on an additional unsupervised segmenter $\Gamma$.
Instead, the attention masks are thresholded. 
We observe that such a version of \NAME{} has strong performance, outperforming prior work as well. 
CutLER is helpful, but not a critical component, and \NAME{} performs strongly without it.

In the same table, we also combine our method with PAMR~\cite{araslanov2020single}, the post-processing approach employed by TCL.
We find that it improves results for our method, though improvements are less drastic since our method already yields better segmentation and boundaries.

Qualitative results are shown in \cref{fig:main_qual}.
This figure highlights a key benefit of our approach: the ability to exploit contextual priors through the use of background prototypes, which in turn allows for the direct assignment of pixels to a background class.
This improves segmentation quality because it makes it easier to differentiate objects from the background and to delineate their boundaries.
In comparison, TCL predictions are very coarse and contain more noise.
\begin{table}[t]
\begin{minipage}[l]{0.31\linewidth}
\centering \footnotesize
\caption{Performance of \NAME{} based on different feature extractors.}
\begin{tabular}{lc}
\toprule
\textbf{Feature} & \multirow{2}{*}{\textbf{VOC}} \\
\textbf{Extractor} & \\
\midrule
MAE & 54.9 \\
DINO & 59.1 \\
CLIP (tokens) & 51.4 \\
CLIP (keys) & 61.8 \\
SD & 64.4 \\
\midrule
SD+CLIP+DINO & 66.4 \\
\bottomrule
\end{tabular}
\label{tab:features}
\end{minipage}\hfill\begin{minipage}[r]{0.66\linewidth}
\centering \footnotesize
\caption{{Ablation of different components.} Each component is removed in isolation, measuring the drop ($\Delta$) in mIoU on VOC and Context datasets.
Using SD features.
}
\label{tab:ablation}
\begin{tabular}{@{}lcccc@{}}
\toprule
\textbf{Configuration} & \textbf{VOC}  & $\mathbf{\Delta}$                     & \textbf{Context} & $\mathbf{\Delta}$                    \\
\midrule

Full                   & 64.4  &                              & 29.4    &                             \\
\midrule
w/o bg prototypes      & 53.2 & {\color[HTML]{FF0000} \footnotesize -11.2} &  28.9   & {\color[HTML]{FF0000} \footnotesize -0.5} \\
w/o category filter                & 54.4 & {\color[HTML]{FF0000} \footnotesize -10.0}  & 25.2    & {\color[HTML]{FF0000} \footnotesize -4.2} \\
w/o ``stuff'' filter     & n/a  & {\color[HTML]{FF0000} \footnotesize }      & 26.9    & {\color[HTML]{FF0000} \footnotesize -2.5} \\
w/o CutLER            & 60.4 & {\color[HTML]{FF0000} \footnotesize -4.0}  & 27.6    & {\color[HTML]{FF0000} \footnotesize -1.8} \\
w/o sliding window                 & 62.2 & {\color[HTML]{FF0000} \footnotesize -2.2}  & 28.6    & {\color[HTML]{FF0000} \footnotesize -0.8} \\
only average $\bar{P}$ & 62.5 & {\color[HTML]{FF0000} \footnotesize -1.9 }  & 28.4    & {\color[HTML]{FF0000} \footnotesize -1.0} \\
\bottomrule
\end{tabular}
\end{minipage}
\end{table}%

\subsection{Ablations}

Next, we ablate the components of \NAME{} on VOC and Context datasets.
For these experiments, only SD is employed as a feature extractor.
We remove individual components and measure the change in segmentation performance, summarising the results in \cref{tab:ablation}.
Our first observation is that background prototypes have a major impact on performance.
When removing them from consideration, we instead threshold the similarity scores of the images with the foreground prototypes (set to 0.72, determined via grid search); in this case, the performance drops significantly, which again highlights the importance of leveraging contextual priors.
\begin{wrapfigure}[9]{r}{0.35\linewidth}
    \centering
    \includegraphics[width=0.3\textwidth,trim={0.8cm 0 0 1cm}]{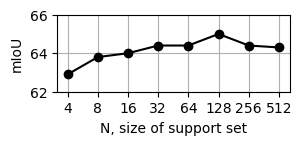}
    \captionof{figure}{PascalVOC results with increasing support size $N$.}
    \label{fig:n_samples}
\end{wrapfigure}
On Context, the impact is less significant, likely due to the fact that the dataset contains ``stuff'' categories. 
Removing the \textit{instance-} and \textit{part-level} prototypes also negatively affects performance.
Additionally, removing the category pre-filtering has a major impact.
We hypothesize that this introduces spurious correlations between prototypes of different classes.
On Context, ``stuff'' filtering is also important. 

We again consider the importance of using an unsupervised segmenter, CutLER, for prototype mask extractions, using thresholding instead.
We find this slightly reduces performance in this setting as well.
Overall, background prototypes and pre-filtering contribute the most.

Finally, we measure the effect of varying the size of the support set $N$ in \cref{fig:n_samples}.
We find that \NAME{} already shows strong performance even at a low number of samples for each query.
With increasing the number of samples, the performance improves, saturating at around $N=32$. which we use in our main experiments.

\subsection{Evaluation without background}
\label{sec:app_nobg}
\begin{table}[t]
\centering \small
\caption{{Comparison with methods when background is excluded (decided by ground truth).} \NAME{} shows comparable performance to prior works despite only relying on pretrained feature extractors. $^*$ result from \cite{cha2022learning}.
}%
\label{tab:app_nobg}
\begin{tabular}{@{}lcccccc@{}}
\toprule
\textbf{Method}       & \textbf{VOC-20}          & \textbf{Context-59}    & \textbf{ADE}  &  \textbf{Stuff} & \textbf{Cityscapes}          \\ 
\midrule
CLIPpy & -- & -- & 13.5 & -- & -- \\
OVSegmentor & -- & -- & 5.6 & -- & --\\
GroupViT$^*$          & \underline{79.7}      & 23.4          & 9.2  & 15.3 & 11.1           \\
MaskCLIP$^*$          & 74.9            & 26.4          & 9.8 & 16.4 & 12.6           \\
ReCo$^*$         & 57.5            & 22.3          & 11.2 & 14.8 & 21.1         \\
TCL          & 77.5            & 30.3          & \textbf{14.9}  & 19.6 & 23.1    \\

\textbf{\NAME{}}         & \textbf{80.9} & \textbf{32.9}  & \underline{14.1} & \textbf{20.3} & \textbf{23.4} \\
 \bottomrule
\end{tabular}
\end{table}

One of the notable advantages of our approach is the ability to represent background regions via (negative) prototypes, leading to improved segmentation performance.
Nevertheless, we hereby also evaluate our method under a different evaluation protocol adopted in prior work, which excludes the \textit{background} class from the evaluation.
We note that prior work often requires additional considerations to handle background, such as thresholding.
In this setting, however, the background class is \textit{not} predicted, and the set of categories, thus, must be exhaustive.
As in practice, this is not the case, and datasets contain unlabelled pixels (or simply a background label), such image areas are removed from consideration.
Consequently, less emphasis is placed  on object boundaries in this setting.
As in this setting the background prediction is invalid, we do not consider negative prototypes.
This setting tests the ability of various methods to discriminate between different classes, which for \NAME{} is inherent to the choice of feature extractors.
Despite this, our method shows competitive performance accross wide range of benchmarks \cref{tab:app_nobg}.

\begin{figure*}[t]
    \centering
    \includegraphics[width=0.92\textwidth]{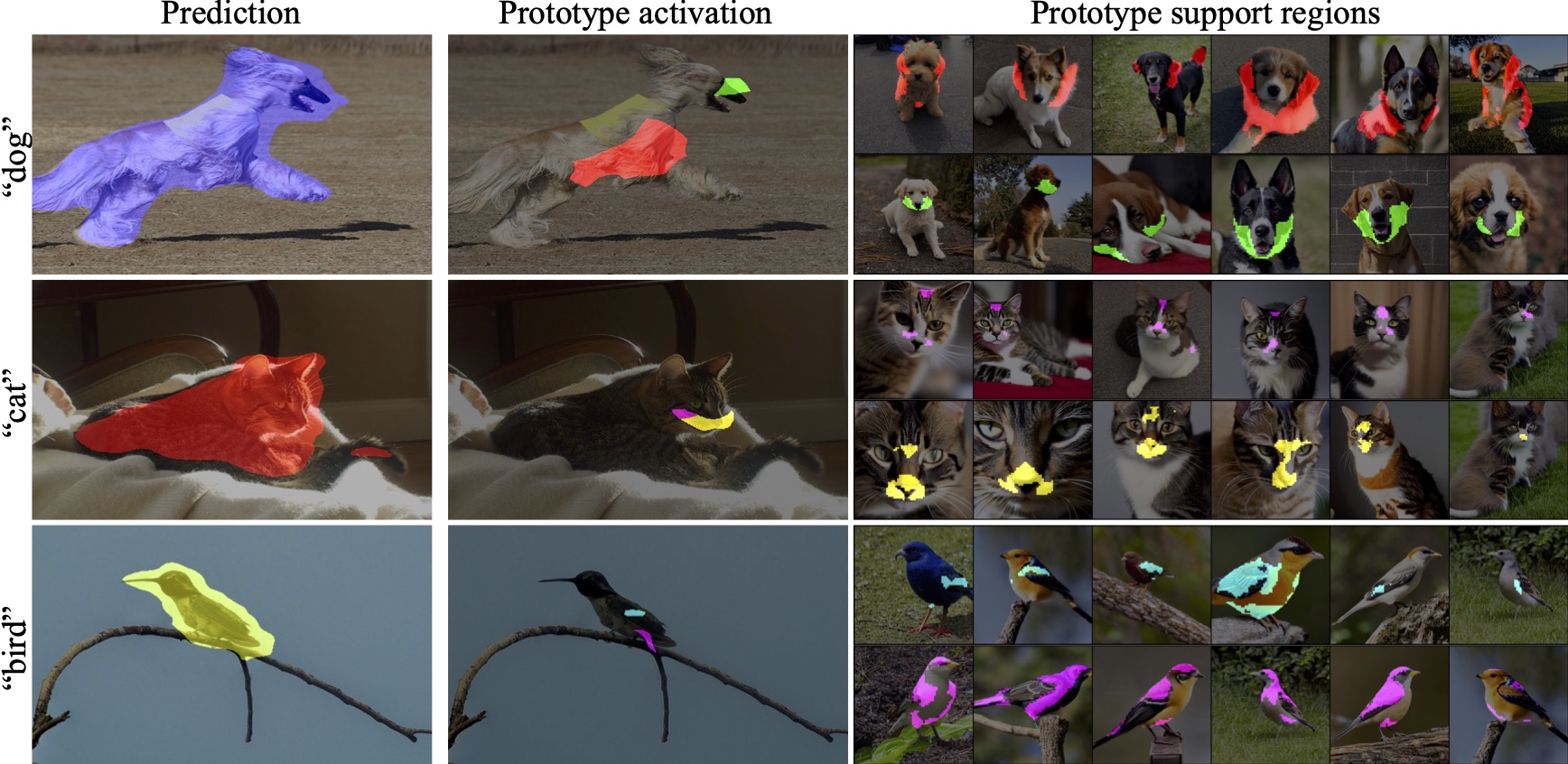}
    \caption{
    Analysis of the segmentation output by linking regions to samples in the support set.
    Left: our results for different classes.
    Middle: select color-coded regions ``activated'' by different prototypes for the class. 
    Right: regions in the support set images corresponding to these (part-level) prototypes.
    }
    \label{fig:explantion}
\end{figure*}%
\subsection{Explaining segmentations}
We inspect how our method segments certain regions by considering which prototype from $\mathcal{P}_{c}^{\operatorname{fg}}$ was used to assign a class $c$ to a pixel.
Prototypes map to regions in the support set from where they were aggregated, \eg, instances prototypes are associated with foreground masks $M^{\operatorname{fg}}_n$ and part prototypes with centroids/clusters.
By following these mappings,
a set of support image regions can be retrieved for each segmentation decision, providing a degree of explainability.
\cref{fig:explantion} illustrates this for examples of \texttt{dog}, \texttt{cat}, and \texttt{bird} classes.
For visualisation purposes, selected prototypes and corresponding regions are shown.
On the left, we show the full segmentation result of each image.
In the middle, we select regions that correlate best with certain class prototypes.
On the right, we retrieve images from the support set and highlight where each prototype emerged.
We find that meaningful part segmentation merges due to clustering the support image features, and similar regions are segmented by corresponding prototypes. 
However, sometimes region covered in the input image will not fully align with the whole prototype (\eg \texttt{cat}'s face around the eyes or lower belly/tail of \texttt{bird}).
Each segmentation is explained by precise regions in a small support set.

\begin{figure*}[t]
    \centering
    \includegraphics[width=0.94\textwidth]{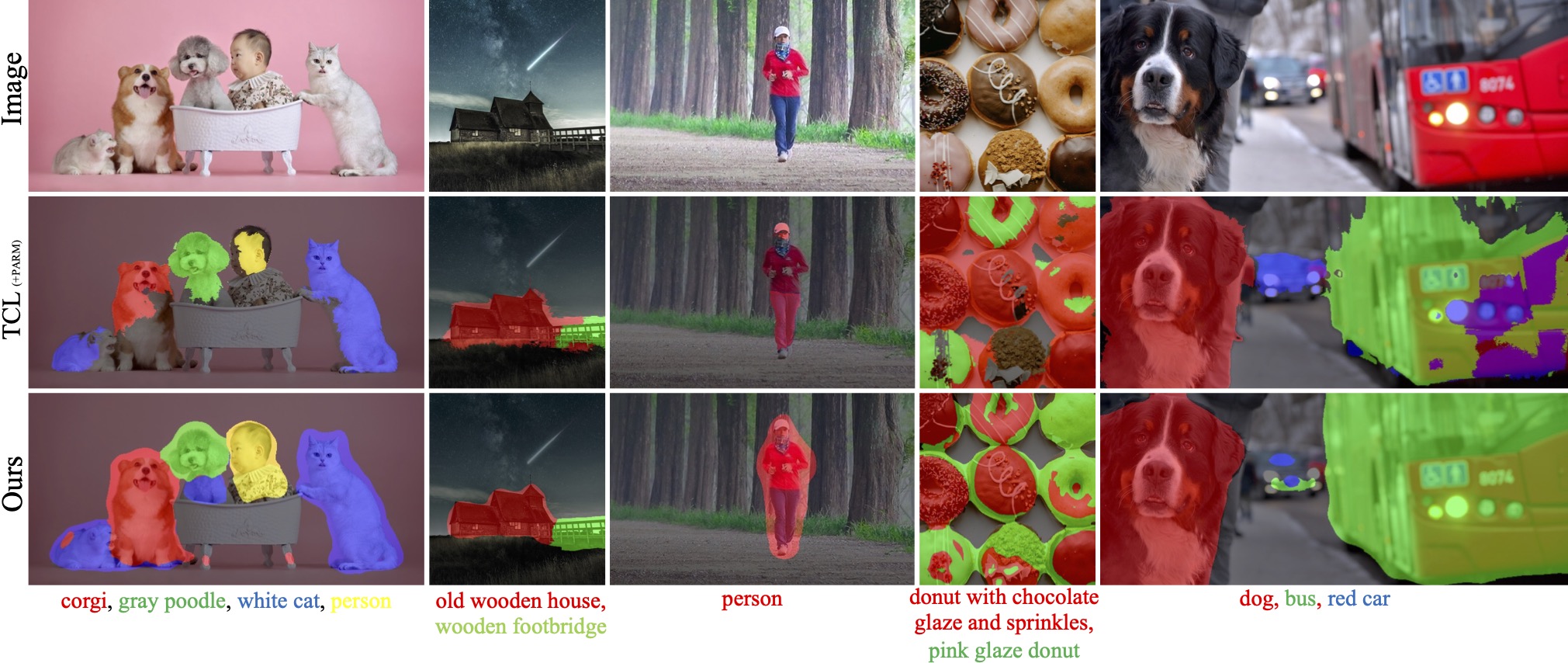}
    \caption{
    Qualitative comparison on challenging in-the-wild images with TCL, which struggles with object boundaries, missing parts of objects, or including surroundings. 
    Our method has more appropriate boundaries and makes fever errors overall, but does produce a small halo effect around objects due to the upscaling of feature extractors.
    }
    \label{fig:in_the_wild}
\end{figure*}
\subsection{In-the-wild}
In \cref{fig:in_the_wild}, we investigate \NAME{} on chal
lenging in-the-wild images with simple and complex backgrounds.
We compare with TCL+PAMR.
In the first three images, both methods correctly detect the objects identified by the queries.
\NAME{} has small false positive "corgi" patches.
TCL however misses large parts of the objects, such as most of the person, and parts of animal bodies.
The distinction between the house and the bridge in the second image is also better with \NAME{}.
We also note that our segmentations sometimes have halos around objects.
This is caused by upscaling the low-resolution feature extractor (SD in this case).
The last two images contain challenging scenarios where both approaches struggle.
The fourth image only contains similar objects of the same type.
Both methods incorrectly identify plain donuts as either of the specified queries.
\NAME{} however correctly identifies chocolate donuts with varied sprinkles and separates all donuts from the background.
In the final picture, the query ``red car'' is added, although no such object is present.
The extra query causes TCL to incorrectly identify parts of the red bus as a car.
Both methods incorrectly segment the gray car in the distance.
However, overall, our method is more robust and delineates objects better despite the lack of specialized training or post-processing.

\section{Conclusion}\label{sec:conclusions}
We introduce \NAME{}, an open-vocabulary segmentation method that operates in two stages. 
First, given queries, support images are sampled and their features are extracted to create class prototypes. 
These prototypes are then compared to features from an inference image.
This approach offers multiple advantages: diverse prototypes accommodating various visual appearances and negative prototypes for background localisation. 
\NAME{} outperforms prior work on benchmarks, exhibiting fewer errors, effectively separating objects from background, and providing explainability through segmentation mapping to support set regions.

\section*{Acknowledgements}

Laurynas Karazija is supported by is supported by AIMS CDT EP/S024050/1.
Iro Laina, Andrea Vedaldi, and Christian Rupprecht are supported by ERC-CoG UNION 101001212 and VisualAI EP/T028572/1.

\paragraph{Ethics.}

For further details on ethics, data protection, and copyright please see \url{https://www.robots.ox.ac.uk/~vedaldi/research/union/ethics.html}.

\bibliographystyle{splncs04}
\bibliography{references}

\clearpage
\appendix
\setcounter{table}{0}
\renewcommand{\thetable}{\Alph{section}.\arabic{table}}
\setcounter{figure}{0}
\renewcommand{\thefigure}{\Alph{section}.\arabic{figure}}
\section*{Supplementary Material}

In this supplementary material, we provide additional experimental results, including further ablations and qualitative comparisons (\cref{sec:app_experiments}), consider the limitations and broader impacts of our work (\cref{sec:app_impact}), 
and conclude with additional details concerning the implementation (\cref{sec:app_impl}). 

\section{Additional experiments}\label{sec:app_experiments}
This section provides additional experimental results of \NAME{}.

\subsection{Additional Comparisons}\label{sec:app_comparisons}
\paragraph{Category filter.} To ensure that the category pre-filtering does not give our approach an unfair advantage, we augment two methods (TCL~\cite{cha2022learning} and OVSegmentor~\cite{xu2023learning}, which are the closest baselines with code and checkpoints available) with our category pre-filtering. We evaluate on the Pascal VOC dataset (where the category filter shows a significant impact; see Table 3) and report the results in \cref{tab:app_cat_filter}. We observe that TCL improves by 0.6, while the performance of OVSegmentor drops by 0.1. On the contrary, our method benefits substantially from this component, but it still shows stronger performance without the filter than baselines with.

\paragraph{Influence of $\Gamma$ segmentation method.}
We also further investigate the use of CutLER~\cite{wang2023cutler} to obtain segmentation masks. We also provide example results of segmentation in \cref{fig:stuff_object}. In \cref{tab:app_cutler}, we devise a baseline where CutLER-predicted masks are used to average the CLIP image encoder's final spatial tokens after projection. Averaged tokens are compared with CLIP text embeddings to assign a class. While relying on pre-trained components (like ours), this avoids support set generation. In the same table, we also consider whether the objectness prior provided by CutLER could be beneficial to other methods as well. 
We consider a version of TCL~\cite{cha2022learning} and OVSegmentor~\cite{xu2023learning} which we augment with CutLER. 
That is, after methods assign class probabilities to each pixel/patch, a majority voting for a class is performed in every region predicted by CutLER.
This combines CutLER's understanding of objects and their boundaries, aspects where prior methods struggle, with open-vocabulary segmentation.
However, we observe that this negatively impacts the performance of these methods, which we attribute to only a limited performance of CutLER in complex scenes present in the datasets.
Finally, we also include a version of \NAME{} that does not rely on CutLER for mask extractions, instead using thresholded masks. We observe that such a version of our method also has strong performance.

We additionally experiment with stronger segmenters to understand the influence of FG/BG mask quality. 
We replace our FG/BG segmentation approach with strong supervised models: with SAM, we achieve 67.1 on VOC, and with Grounded SAM, 68.5. This slightly improves results from 66.3 of our configuration with CutLER, but the performance gain is not large and thus not critical.

\begin{wraptable}[11]{r}{0.25\linewidth}
    \vspace{-9pt}
    \centering
    \caption{Influence of different text-to-image generators.}
    \label{tab:model_values}
    \begin{tabular}{@{}cc@{}}
        \toprule
        \textbf{T2I} & \textbf{VOC} \\
        \midrule
        SD 1.5 & 66.4 \\
        SD 2.0 & 67.7 \\
        SD 2.1 & 67.1 \\
        Hyper-SD & 67.7 \\
        \bottomrule
    \end{tabular}
\end{wraptable}%
\paragraph{Influence of image generator.}
We experiment with different SD versions in \cref{tab:model_values} and observe improvement with more advanced generators.

\paragraph{Class prompts.} We additionally consider whether corrections introduced to class prompts might have similarly provided additional benefits to our approach (see \cref{sec:app_datasets} for details). To that end, we also evaluate TCL and OVSegmenter (methods that do not rely on additional prompt curation) with our corrected prompts and consider a version of our method without such corrections in \cref{tab:app_prompts}. We observe only marginal to no impact on the performance.
\paragraph{Prompt template} Finally, we consider the prompt template employed when sampling support image set: ``\texttt{A good picture of a $\langle c_i \rangle$}'' for class prompt $c_i$. This template is generic and broadly applicable to virtually any natural language specification of a target class. While prior work adopts prompt expansion by considering a list of synonyms and subcategories, it is not entirely clear how such a strategy could be systematically performed for any in-the-wild prompts, such as a “chocolate glazed donut”. We experiment with a list of synonyms and subclasses, as employed by~\cite{ranasinghe2022perceptual}, on VOC datasets measuring 66.4 mIoU, which is similar to our single prompt performance $66.3\pm0.2$. Curating such lists automatically is an interesting future scaling direction.

\subsection{Additional ablations}\label{sec:app_ablations}
\begin{table}[t]
\centering \small
\caption{{Use of category filter component}. \NAME{} without category filter outperforms prior work with cat. filter.}
\label{tab:app_cat_filter}
\begin{tabular}{@{}lcc@{}}
\toprule
\multirow{2}{*}{\textbf{Model}}   & \multicolumn{2}{l}{\textbf{Category filter}}                         \\
                         & \ding{55}        & $\checkmark$         \\ \midrule
OVSegmentor              & 53.8                         & 53.7                         \\
TCL                      & 51.2                         & 51.8                         \\
TCL (+PAMR)              & 55.0                         & 56.0                         \\
\NAME{} & \textbf{56.2} & \textbf{66.4} \\
\bottomrule
\end{tabular}
\centering \small
\caption{{Application of CutLER}. Prior work does not benefit from using CutLER during inference, while \NAME{} shows strong results without it.}
\label{tab:app_cutler}
\begin{tabular}{@{}l@{\hspace{-1pt}}c@{\hspace{-1pt}}ccc@{}}
\toprule
\textbf{Model}       & \textbf{CutLER}    & \textbf{VOC}          & \textbf{Context}      & \textbf{Object}       \\ \midrule
CLIP        & \checkmark & 33.0         & 11.6         & 11.1         \\
OVSegmentor &            & 53.8         & 20.4         & 25.1         \\
OVSegmentor & \checkmark & 38.7         & 14.4         & 16.8         \\
TCL         &            & 51.2         & 24.3         & 30.4         \\
TCL         & \checkmark & 43.1         & 20.5         & 22.7         \\
OVDiff      &           & 62.8 & 28.6 & 34.9 \\
OVDiff      & \checkmark & \textbf{66.3 $\pm$ 0.2} & \textbf{29.7 $\pm$ 0.3} &  \textbf{34.6 $\pm$ 0.3} \\
\bottomrule
\end{tabular}
\end{table}
\begin{table}[t]
\centering \small
\caption{{Using corrected prompts}. We consider if corrected class names benefit prior work. We observe negligible to no effect.}
\label{tab:app_prompts}
\begin{tabular}{@{}l@{\hspace{-5pt}}c@{\hspace{2pt}}ccc@{}}
\toprule
\textbf{Model}       & \textbf{Correction}    & \textbf{VOC}          & \textbf{Context}      & \textbf{Object}       \\ \midrule
OVSegmentor &            & 53.8         & 20.4         & 25.1         \\
OVSegmentor & \checkmark & 53.9         & 20.4         & 25.1         \\
TCL         &            & 51.2         & 24.3         & 30.4         \\
TCL         & \checkmark & 50.6         & 24.3         & 30.4         \\
OVDiff      &            & 66.1 & 29.5 & 34.9  \\
OVDiff      & \checkmark & \textbf{66.3 $\pm$ 0.2} & \textbf{29.7 $\pm$ 0.3}  &  \textbf{34.6 $\pm$ 0.3} \\
\bottomrule
\end{tabular}
\centering \small
\caption{{Choice of $K$ for number of centroids}.}
\label{tab:app_k}
\begin{tabular}{ccc}
\toprule
$\mathbf{K}$  & \textbf{VOC} & \textbf{Context} \\
\midrule
8  & 63.8 & 29.2    \\
16 & 64.0 & 29.3    \\
32 & 64.4 & 29.4    \\
64 & 64.3 & 28.0    \\
\bottomrule
\end{tabular}
\end{table}
\begin{table}[t]
\centering \small
\caption{{Ablation of different SD feature configurations.} Removing first and last cross attention \textit{layers}, mid, 1$^\mathrm{st}$ and 2$^\mathrm{nd}$ upsampling \textit{blocks} (all layers in the block) has a negative effect.}
\begin{tabular}{@{}cccccc@{}}
\toprule
\textbf{1st} & \textbf{Mid} & \textbf{Up-1} & \textbf{Up-2} & \textbf{Last} &  \\
\textbf{layer} & \textbf{block} & \textbf{block} & \textbf{block} & \textbf{layer}  & \textbf{Context} \\
\midrule
$\checkmark$ & $\checkmark$ & $\checkmark$ & $\checkmark$ & $\checkmark$ & 29.4    \\
             & $\checkmark$ & $\checkmark$ & $\checkmark$ & $\checkmark$ & 29.4    \\
$\checkmark$ &              & $\checkmark$ & $\checkmark$ & $\checkmark$ & 29.2    \\
$\checkmark$ & $\checkmark$ &              & $\checkmark$ & $\checkmark$ & 27.3    \\
$\checkmark$ & $\checkmark$ & $\checkmark$ &              & $\checkmark$ & 28.9    \\
$\checkmark$ & $\checkmark$ & $\checkmark$ & $\checkmark$ &              & 29.3    \\
\bottomrule
\end{tabular}
\label{tab:app_feature_ablation}

\end{table}
\begin{table}[t]
\centering \small
\caption{{Ablation of various configurations for prototypes.} We consider average $\bar{P}$, instance $P_n$, and part $P_k$ prototypes individually and in various combinations on VOC and Context datasets. Combination of all three types of prototypes shows strongest results.}
\label{tab:app_proto_ablation}
\begin{tabular}{ccccc}
\toprule
$\mathbf{\bar{P}}$ & $\mathbf{P_n}$ & $\mathbf{P_k}$ & \textbf{VOC}  & \textbf{Context} \\
\midrule
$\checkmark$ & $\checkmark$    & $\checkmark$      & 64.4 & 29.4    \\
$\checkmark$ &                 & $\checkmark$      & 61.7 & 29.3    \\
$\checkmark$ & $\checkmark$    &                   & 63.5 & 29.4    \\
             & $\checkmark$    & $\checkmark$      & 62.5 & 28.4     \\
             &                 & $\checkmark$      & 63.7 & 28.8    \\
             & $\checkmark$    &                   & 60.0 & 29.0    \\
$\checkmark$ &                 &                   & 62.5 & 28.4    \\
\bottomrule
\end{tabular}
\end{table}
\paragraph{Prototype combinations.} In \cref{tab:app_proto_ablation}, we consider the three different types of prototypes described in 
Section 3 and test their performance individually and in various combinations. 
We find that the ``part'' prototypes obtained by $K$-means clustering show strong performance when considered individually on VOC. Instance prototypes show strong individual performance on Context, as well as in combination with the average category prototype. 
The combination of all three types shows the strongest results across the two datasets, which is what we adopt in our main set of experiments.

We also consider the treatment of prototypes under the stuff filter. We investigate the impact of not excluding background prototypes for ``stuff" classes. In this setting, we measure 29.1 on Context, which is a slight reduction in performance. We also investigate the benefit of categorisation into ``things'' and ``stuff'' used in the stuff filter component. Instead, we filter all background prototypes using all foreground prototypes. In this configuration, we measure 27.6 on Context. Both configurations show a reduction from 29.4, measuring using the stuff filter with categorisation in ``stuff'' and ``things'', as used in our main experiments. Finally, we experiment by removing part-level prototypes for ``stuff'' classes, which also results in a performance drop to 28.0.

\paragraph{$K$ - number of clusters.} In \cref{tab:app_k}, we investigate the sensitivity of the method to the choice of $K$ for the number of ``part'' prototypes extracted using $K$-means clustering. Although our setting $K=32$ obtains slightly better results on  Context and VOC, other values result in comparable segmentation performance suggesting that \NAME{} is not sensitive to the choice of $K$ and a range of values is viable.

\begin{figure*}[t]
    \centering
    \includegraphics[width=0.95\textwidth]{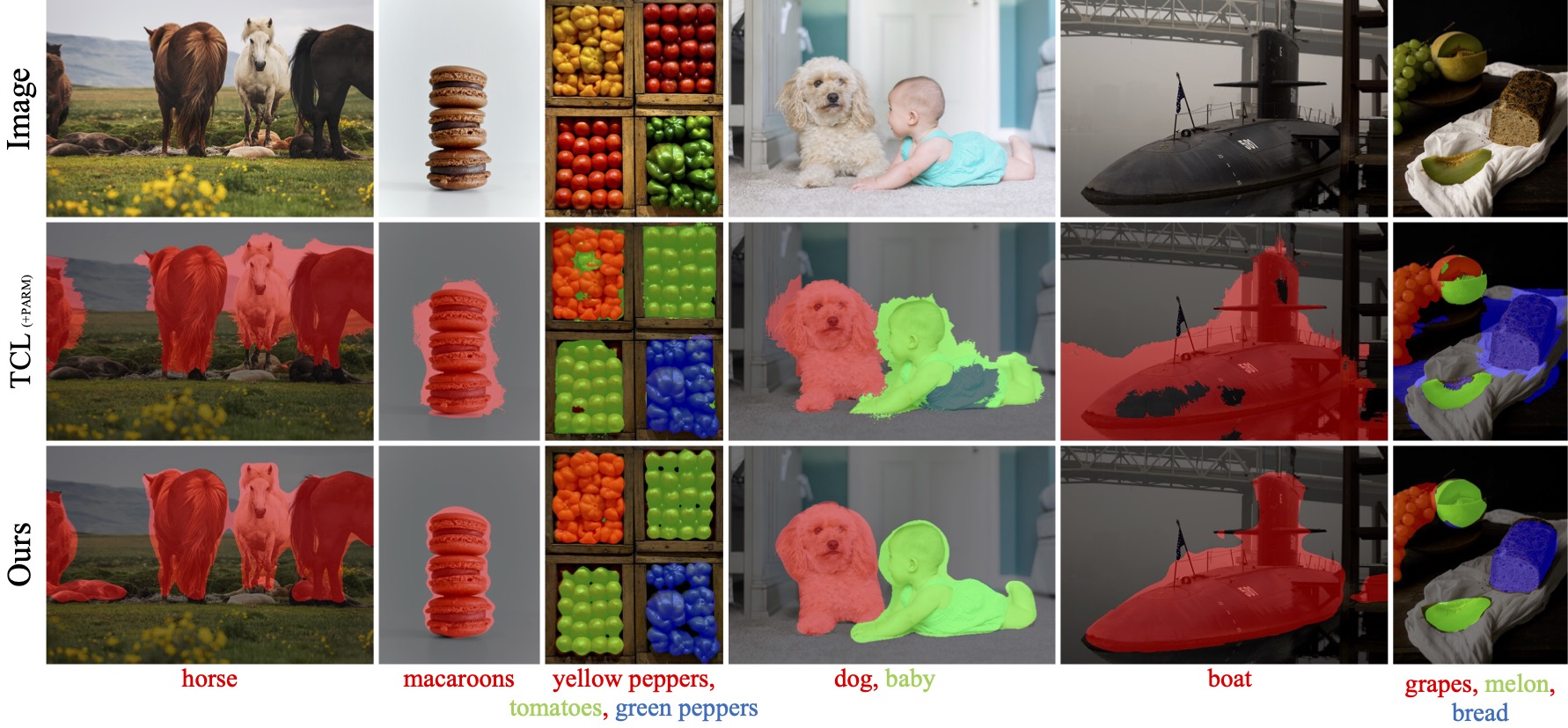}
    \caption{{Qualitative comparison on in-the-wild images.} \NAME{} performs significantly better than prior state-of-the-art, TCL, on wildlife images containing multiple instances, studio photos with simple backgrounds, images containing multiple categories and an image containing a rare instance of a class.
    }
    \label{fig:app_in_the_wild}
\end{figure*}

\paragraph{SD features.} 
When using Stable Diffusion as a feature extractor, we consider various combinations of layers/blocks in the UNet architecture. 
We follow the nomenclature used in the Stable Diffusion implementation where consecutive layers of Unet are organised into \textit{blocks}. There are 3 down-sampling blocks with 2 cross-attention layers each, a mid-block with a single cross-attention, and 3 up-sampling blocks with 3 cross-attention layers each. 
We report our findings in \cref{tab:app_feature_ablation}.
Including the first and last cross-attention layers in the feature extraction process has a small positive impact on segmentation performance, which we attribute to the high feature resolution.
We also consider excluding features from the middle block of the network due to small $8\times8$ resolution  but observe a small negative impact on performance on the Context dataset.
We also investigate whether including the first (Up-1) and the second upsampling (Up-2) blocks are necessary.
Without them, the performance drops the most out of the configurations considered.
Thus, we use a concatenation of features from the middle, first and second upsampling blocks and the first and last layers in our main experiments.

\subsection{Qualitative results}
\begin{figure*}[t]
    \centering
    \includegraphics[width=\textwidth]{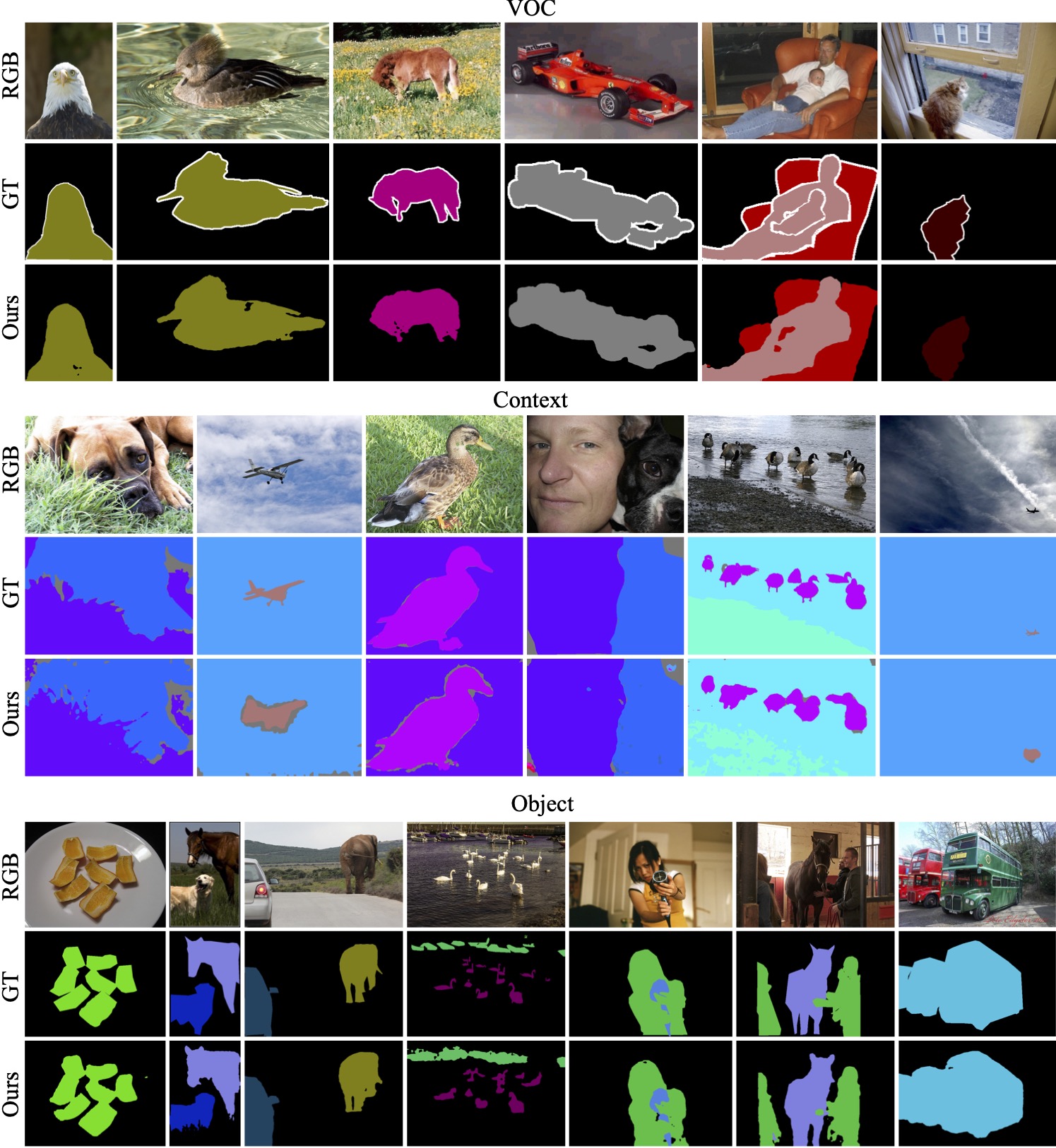}
    \caption{{Additional qualitative results}. Images from Pascal VOC (top), Pascal Context (middle), and COCO Object (bottom).}
    \label{fig:app_qual_fig}
\end{figure*}

\begin{figure*}
    \begin{subfigure}[l]{0.49\textwidth}
        \includegraphics[width=\textwidth]{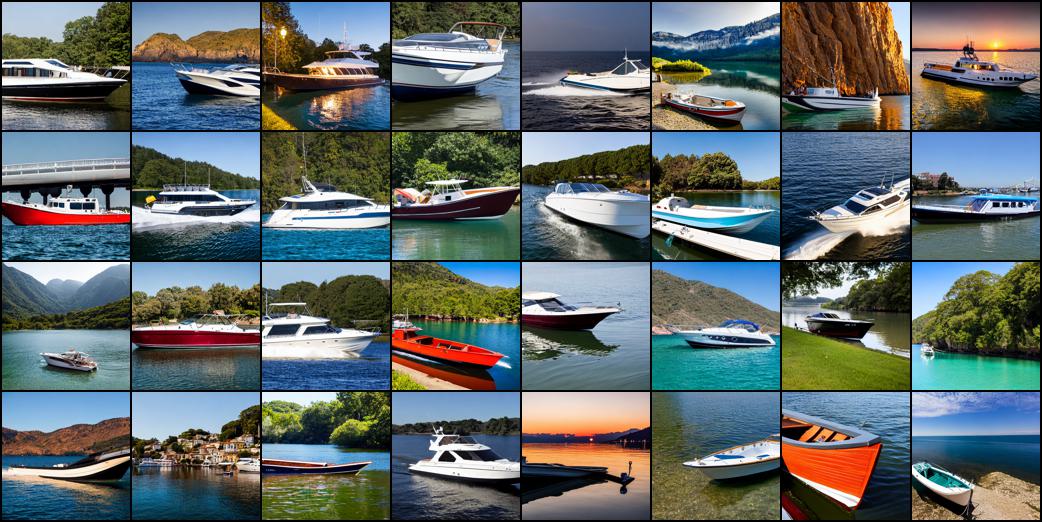}
        \caption{boat}
    \end{subfigure}
    \hfill
    \begin{subfigure}[r]{0.49\textwidth}
        \includegraphics[width=\textwidth]{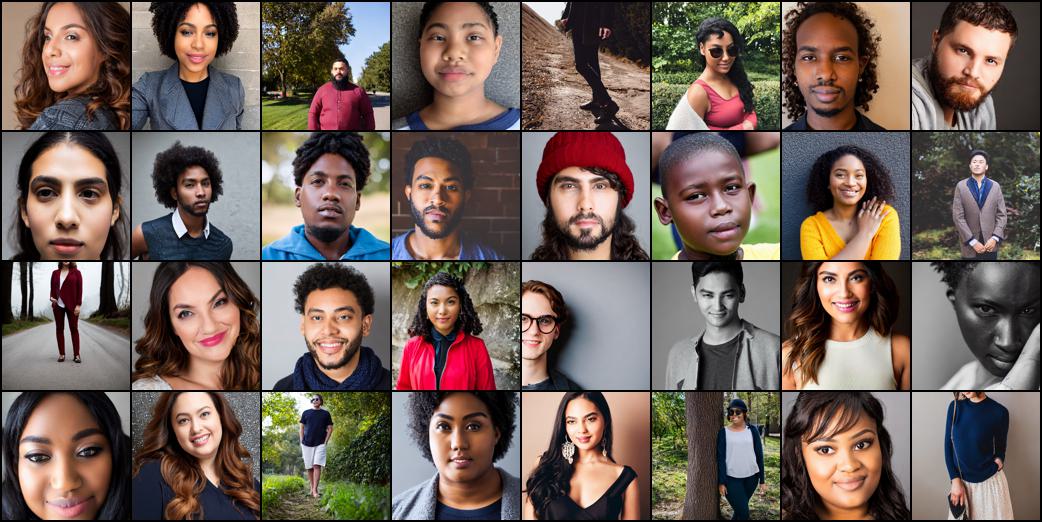}
        \caption{person}
    \end{subfigure}
    \begin{subfigure}[l]{0.49\textwidth}
        \includegraphics[width=\textwidth]{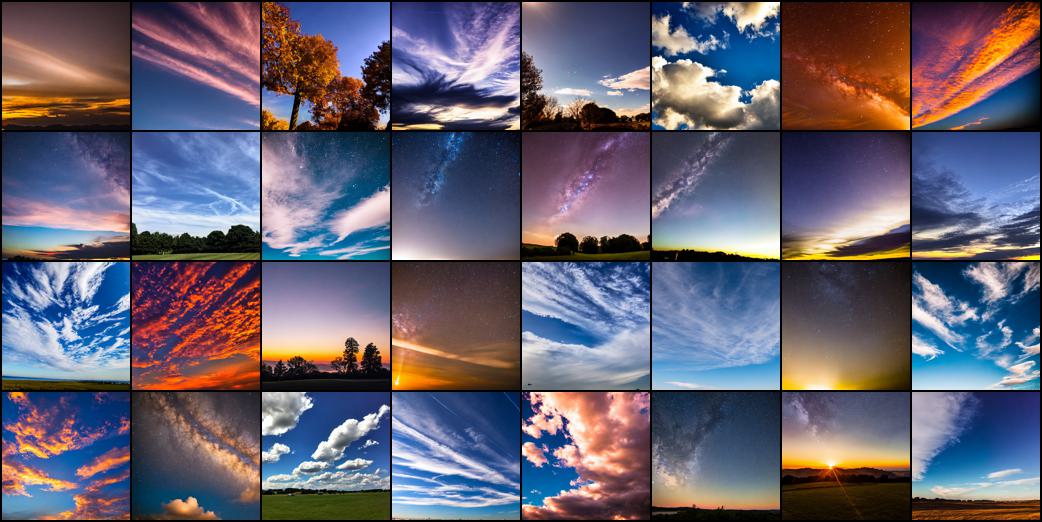}
        \caption{sky}
    \end{subfigure}
    \hfill
    \begin{subfigure}[r]{0.49\textwidth}
        \includegraphics[width=\textwidth]{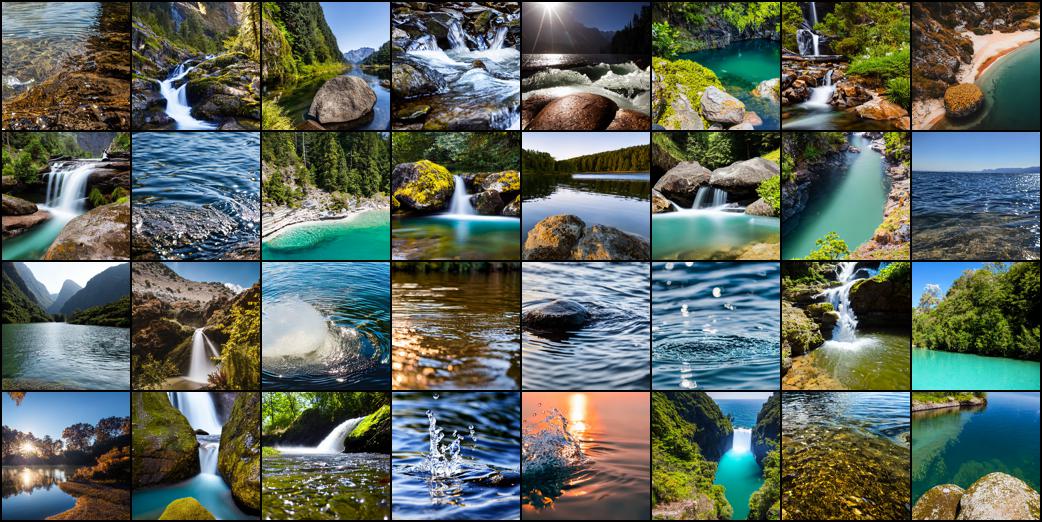}
        \caption{water}
    \end{subfigure}
    \begin{subfigure}[l]{0.49\textwidth}
        \includegraphics[width=\textwidth]{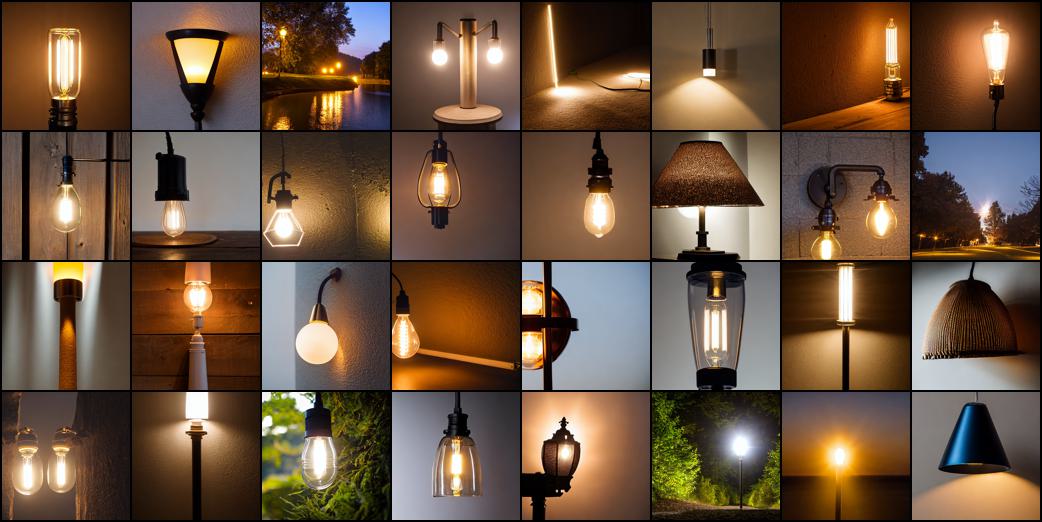}
        \caption{light}
    \end{subfigure}
    \hfill
    \begin{subfigure}[r]{0.49\textwidth}
        \includegraphics[width=\textwidth]{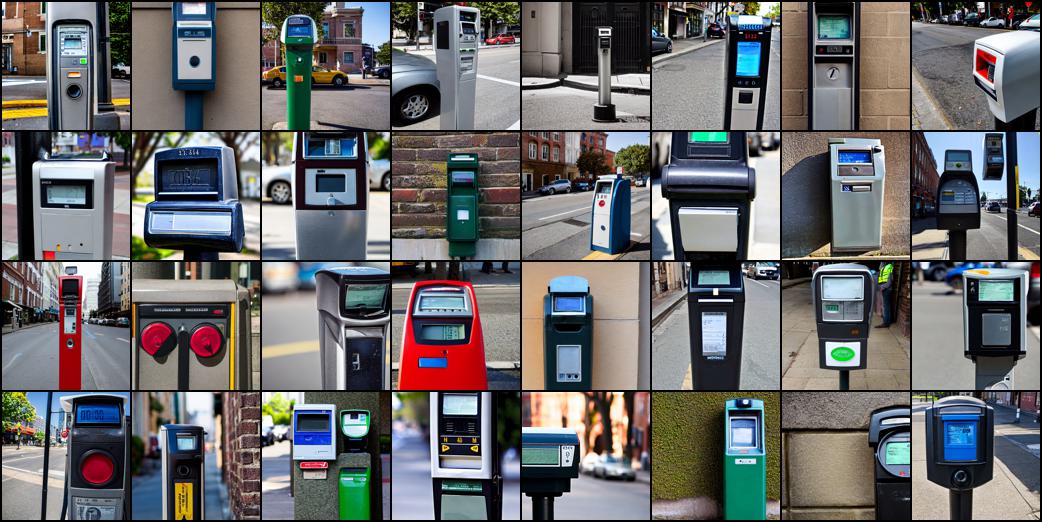}
        \caption{parking meter}
    \end{subfigure}
    \hfill
    \begin{subfigure}[r]{0.49\textwidth}
        \includegraphics[width=\textwidth]{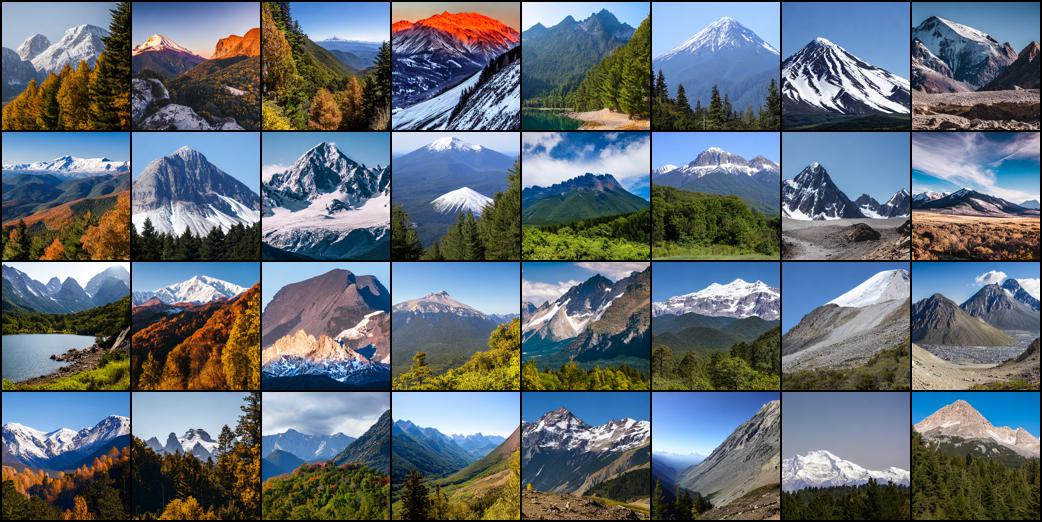}
        \caption{mountain}
    \end{subfigure}
    \hfill
    \begin{subfigure}[r]{0.49\textwidth}
        \includegraphics[width=\textwidth]{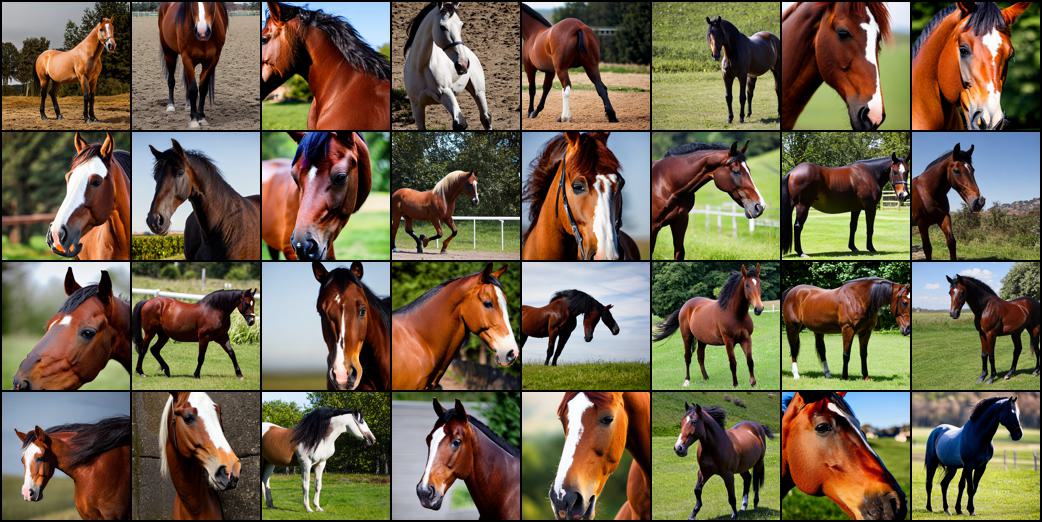}
        \caption{horse}
    \end{subfigure}
    \caption{
    {Images sampled for a support set of some categories.}
    }
    \label{fig:app_classes}
\end{figure*}

\begin{figure}[]
\begin{minipage}{0.49\linewidth}
        \centering
        \includegraphics[width=0.75\linewidth]{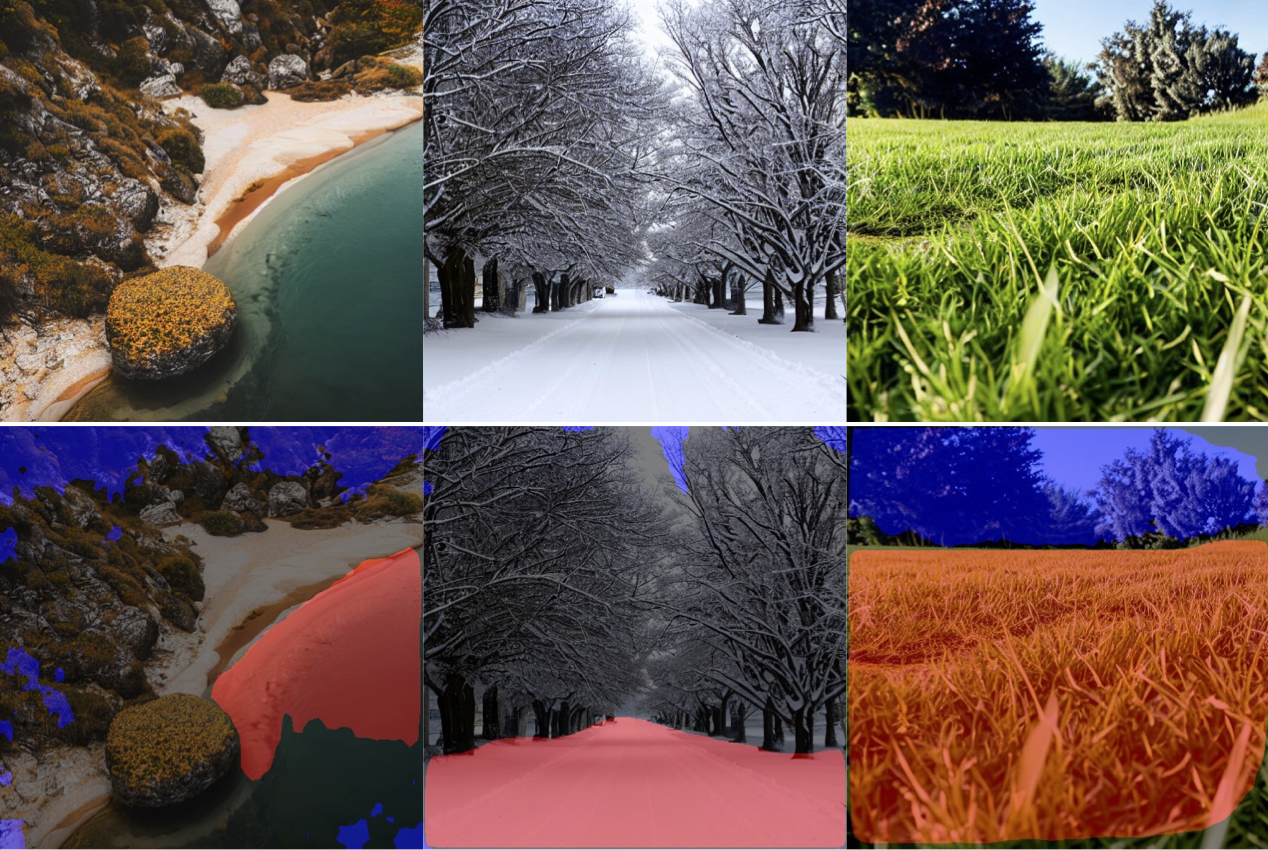}
        \caption{FG/BG segmentation of classes of \textit{water}, \textit{snow} and \textit{grass}. The foreground is in red, while the background is shown in blue.}
        \label{fig:stuff_object}
    \end{minipage}
    \hfill
    \begin{minipage}{0.49\linewidth}
        \centering
        \includegraphics[width=0.75\linewidth]{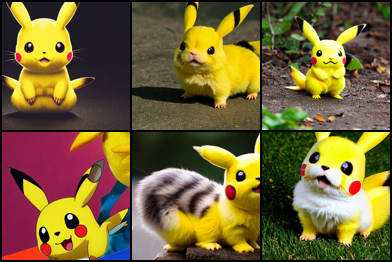}
        \caption{Example images from the support set of a rare \textit{pikachu} class.}
        \label{fig:pika}
    \end{minipage}
\end{figure}

We include additional qualitative results from the benchmark datasets in \cref{fig:app_qual_fig}.
Our method achieves high-quality segmentation across all examples without any post-processing or refinement steps.
In \cref{fig:app_classes}, we show examples of support images sampled for some things, and stuff categories.
In \cref{fig:pika}, we show examples of support set images sampled for rare \textit{pikachu} class. 


\section{Broader impact}\label{sec:app_impact}
Semantic segmentation is a component in a vast and diverse spectrum of applications in healthcare, image processing, computer graphics, surveillance and more.
As for any foundational technology, applications can be good or bad.
\NAME{} is similarly widely applicable.
It also makes it easier to use semantic segmentation in new applications by leveraging existing and new pre-trained models.
This is a bonus for inclusivity, affordability, and, potentially, environmental impact (as it requires no additional training, which is usually computationally intensive); however, these features also mean that it is easier for bad actors to use the technology.

Because \NAME{} does not require further training, it is more versatile but also inherits the weaknesses of the components it is built on. For example, it might contain the biases (e.g., gender bias) of its components, in particular Stable Diffusion~\cite{Schramowski_2023_CVPR}, which is used for generating support images for any given category/description. Thus, it should not be exposed without further filtering and detection of, e.g., NSFW material in the sampled support set. Finally, \NAME{} is also bound by the licenses of its components.

\subsection{Limitations}\label{sec:limitations}
As \NAME{} relies on pretrained components, it inherits some of their limitations.
\NAME{} works with the limited resolution of feature extractors, due to which it might occasionally miss tiny objects.
Furthermore, \NAME{} cannot segment what the generator cannot generate. 
For example, current diffusion models struggle with producing legible text, which can make it difficult to segment specific words.
Furthermore, applications in domains far from the generator's training data (\eg medical imaging) are unlikely to work out of the box.

\section{\NAME{}: Further details}\label{sec:app_impl}

In this section, we provide additional details concerning the implementation of \NAME{}. We begin with a brief overview of the attention mechanism and diffusion models central to extracting features and sampling images.
We review different feature extractors used. We specify the hyperparameter setting for all our experiments and provide an overview of the exchange with ChatGPT used to categorise classes into ``thing'' and ``stuff''. 

\subsection{Preliminaries}\label{sec:app_prelims}

\paragraph{Attention.}
In this work, we make use of pre-trained ViT~\cite{dosovitskiyimage} networks as feature extractors, which repeatedly apply multi-headed attention layers.
In an attention layer, input sequences $X \in \mathbb{R}^{l_x \times d}$ and $Y \in \mathbb{R}^{l_y \times d}$ are linearly project to forms \textit{keys}, \textit{queries}, and \textit{values}: $K = W_kY,\; Q=W_qX,\; V=W_vX$.
In self-attention, $X=Y$.
Attention is calculated as $A = \mathrm{softmax}(\frac{1}{\sqrt d}QK^\top)$, and softmax is applied along the sequence dimension $l_y$.  %
The layer outputs an update $Z=X + A\cdot V$.
ViTs use multiple heads, replicating the above process in parallel with different projection matrices $W_k,W_q,W_v$.
In this work, we consider \textit{queries} and \textit{keys} of attention layers as points where useful features that form meaningful inner products can be extracted.
As we detail later (\Cref{sec:app_feature_extractors}), we use the \textit{keys} from attention layers of ViT feature extractors (DINO/MAE/CLIP), concatenating multiple heads if present.

\paragraph{Text-to-image diffusion models.}
Diffusion models are a class of generative models that form samples starting with noise and gradually denoising it.
We focus on latent diffusion models~\cite{rombach2022high} which operate in the latent space of an image VAE~\cite{kingma2014auto} forming powerful conditional image generators.
During training, an image is encoded into VAE latent space, forming a latent vector $z_0$.
A noise is injected forming a sample
$z_\tau \sim \mathcal{N}(z_\tau; \sqrt{1-\alpha_{\tau}}z_0,\alpha_\tau I)$ for timestep $\tau \in \{1\dots T\}$, where $\alpha_\tau$ are variance values that define a noise schedule such that the resulting $z_T$ is approximately unit normal.
A conditional UNet~\cite{ronneberger2015u}, $\epsilon_\theta(z_t, t, c)$, is trained to predict the injected noise, minimising the mean squared error $\mathbb{E}_t \left(\alpha_t\|\epsilon_\theta(z_t, t, c) - z_0\|_2\right)$ for some caption $c$ and additional constants $a_t$.
The network forms new samples by reversing the noise-injecting chain. Starting from $\hat{z}_T \sim \mathcal{N}(\hat{z}_T;0,I)$, one iterates $\hat{z}_{t-1} = \frac{1}{\sqrt{1-\alpha_t}}(\hat{z}_{t} + \alpha_t\epsilon_\theta(\hat{z}_t, t, c)) + \sqrt{\alpha_t}\hat{z_t}$ until $\hat{z}_0$ is formed and decoded into image space using the VAE decoder.
The conditional UNet uses cross-attention layers between image patches and language (CLIP) embeddings to condition on text $c$ and achieve text-to-image generation.

\subsection{Feature extractors}\label{sec:app_feature_extractors}

\NAME{} is buildable on top of any pre-trained feature extractor. 
In our experiments, we have considered several networks as feature extractors with various self-supervised training regimes:
\begin{itemize}
    \item \textbf{DINO}~\cite{caron2021emerging} is a self-supervised method that trains networks by exploring alignment between multiple views using an  exponential moving average teacher network. We use the ViT-B/8 model pre-trained on ImageNet\footnote{Model and code available at \url{https://github.com/facebookresearch/dino}.} and extract features from the \textit{keys} of the last attention layer.
    \item \textbf{MAE}~\cite{he2017mask} is a self-supervised method that uses masked image inpainting as a learning objective, where a portion of image patches are dropped, and the network seeks to reconstruct the full input. We use the ViT-L/16 model pre-trained on ImageNet at a resolution of 448~\cite{hu2022exploring}.\footnote{Model and code from \url{https://github.com/facebookresearch/long_seq_mae}.} The \textit{keys} of the last layer of the \textit{encoder} network are used. No masking is performed.
    \item \textbf{CLIP}~\cite{radford2021learning} is trained using image-text pairs on an internal dataset WIT-400M. We use ViT-B/16 model\footnote{Model and code from \url{https://github.com/openai/CLIP}.}. We consider two locations to obtain dense features: \textit{keys} from a self-attention layer of the image encoder and \textit{tokens} which are the outputs of transformer layers. We find that \textit{keys} of the second-to-last layer give better performance.

    \item We also consider \textbf{Stable Diffusion}\footnote{We use implementation from \url{https://github.com/huggingface/diffusers}.} (v1.5) itself as a feature extractor. To that end, we use the \textit{queries} from the cross-attention layers in the UNet denoiser, which correspond to the image modality.
    Its UNet is organised into three downsampling blocks, a middle block, and three upsampling blocks. 
    We observe that the middle layers have the most semantic content, so we consider the middle block, 1st and 2nd upsampling blocks and aggregate features from all three cross-attention layers in each block. As the features are quite low in resolution, we include the first downsampling cross-attention layer and the last upsampling cross-attention layer as well. The feature maps are bilinearly upsampled to resolution $64\times64$ and concatenated. A noise appropriate for $\tau=200$ timesteps is added to the input. For feature extraction, we run SD in \textit{unconditional} mode, supplying an empty string for text caption. 
\end{itemize}

\subsection{Datasets}\label{sec:app_datasets}
We evaluate on validation splits of PASCAL VOC (VOC), Pascal Context (Context) and COCO-Object (Object) datasets. PASCAL VOC~\cite{everingham2010pascal,pascal-voc-2012} has 21 classes: 20 foreground plus a background class. For Pascal Context~\cite{mottaghi2014role}, we use the common variant with 59 foreground classes and 1 background class. It contains both “things” and “stuff” classes. The COCO-Object is a variant of COCO-Stuff~\cite{caesar2018cocostuff} with 80 “thing” classes and one class for the background. Textual class names are used as natural language specifications of names.
We renamed or specified certain class names to fix errors (\eg \texttt{pottedplant} $\rightarrow$ \texttt{potted plant}), resolve ambiguity better (\eg \texttt{mouse} $\rightarrow$ \texttt{computer mouse}) or change to more common spelling/word (\eg \texttt{aeroplane} $\rightarrow$ \texttt{airplane}), resulting in 14 fixes. We experiment and measure the impact of this in \cref{sec:app_comparisons} for our and prior work.

\subsection{Comparative baselines}\label{sec:app_baselines}
We briefly review the prior work in used in our experiments, mainly in Table 1.
We consider baselines that do not rely on mask annotations and have code and checkpoints available or detail their evaluation protocol that matches that used in other prior works~\cite{xu2022groupvit,cha2022learning,xu2023learning}.%
Most prior work~\cite{liu2022open,cha2022learning,xu2022groupvit,ren2023viewco,lou2022segclip,xu2023learning} trains image and text encoders on large image-text datasets with a contrastive loss. The methods mainly differ in their architecture and use of grouping mechanisms to ground image-level text on regions. ViL-Seg~\cite{liu2022open} uses online clustering, GroupViT~\cite{xu2022groupvit} and ViewCo~\cite{ren2023viewco} employ group tokens. OVSegmentor~\cite{xu2023learning} uses slot-attention and SegCLIP~\cite{lou2022segclip} a grouping mechanism with learnable centers. CLIPPy~\cite{ranasinghe2022perceptual}, TCL~\cite{cha2022learning}, and MaskCLIP~\cite{zhou2022maskclip} predict classes for each image patch:~\cite{ranasinghe2022perceptual} use max-pooling aggregation,~\cite{cha2022learning} self-masking, and~\cite{zhou2022maskclip} modify CLIP for dense predictions. To assign a background label~\cite{liu2022open,cha2022learning,xu2022groupvit,ren2023viewco,lou2022segclip} use thresholding while~
\cite{ranasinghe2022perceptual} uses dataset-specific prompts. 
CLIP-DIY~\cite{wysoczanska2024clip} leverages CLIP as a zero-shot classifier and applies it on multiple scales to form a dense segmentation. 
ReCO~\cite{shin2022reco} is closer in spirit to our approach as it uses a support set for each prompt; this set, however, is CLIP-retrieved from curated image collections, which may not be applicable for any category in-the-wild. The conceptual difference between OVDiff and ReCO is that OVDiff emphasises and preserves \emph{diverse} prototypes by construction: generation overcomes a limited database; sampled images are segmented individually preserving unique visuals of each instance rather than co-segmenting, which leverages commonality. 
We construct multiple prototypes at multiple levels of granularity to similar effect, as opposed to averaging in ReCO. 

We also note that prior work builds on top of similar pre-trained components such as CLIP in~\cite{shin2022reco,zhou2022maskclip,cha2022learning,lou2022segclip}, OpenCLIP in \cite{wysoczanska2024clip}, DINO + T5/RoBERTa in~\cite{ranasinghe2022perceptual,xu2023learning}. We additionally make use of StableDiffusion, which is trained on a larger dataset (3B, compared to 400M of CLIP or 2B or OpenCLIP).
\NAME{} is, however, fundamentally different to all prior work, as (a) it generates a support set of synthetic images given a class description, and (b) it does not rely on additional training data and further training for learning to segment.

\subsection{Hyperparameters}\label{sec:app_hyperparameters}
\NAME{} has relatively few hyperparameters and we use the same set in all experiments.
Unless otherwise specified, $N=32$ images are sampled using classifier-free guidance scale~\cite{ho2021classifier} of 8.0 and 30 denoising steps. We employ \texttt{DPM-Solver} scheduler~\cite{lu2022dpm}. When sampling images for the support sets, we also use a negative prompt ``\textit{text, low quality, blurry, cartoon, meme, low resolution, bad, poor, faded}". If/when segmenter $\Gamma$ fails to extract any components in a sampled image, a fallback of
adaptive thresholding of $A_n$ is used, following \cite{liao2001fast}.
During inference, we set $\eta = 10$, which results in 1024 text prompts processed in parallel, a choice made mainly due to computational constraints. We set the thresholds for the ``stuff'' filter between background prototypes for ``things'' classes and the foreground of ``stuff'' at 0.85 for all feature extractors. When sampling, a seed is set for each category individually to aid reproducibility. 

\paragraph{Computation cost.} 
We focus on a construction of a method to show that existing foundational diffusion models can be used for segmentation with great efficacy without further training.
\NAME{} requires computing prototypes instead.
With our unoptimized implementation, we measure around $110\pm10$s to calculate prototypes (sample images, extract features and aggregate) for a single category or $50.2\pm2$s without clustering using SD. Using CLIP, we measure  $49.2\pm0.2$s with clustering and $47.7\pm0.2$s without. We note that sampling time grows linearly: we measure 55s for 16, 110s for 32, and 213s for 64 images per class.
The prototype storage requirements are 
0.39MB using CLIP/DINO for each class.

With our unoptimized implementation, we measure around $110\pm10$s to calculate prototypes using SD for a single class, or around 1.14 TFLOP/s-hours of compute. 
While the focus of this study is not computational efficiency, we can compare prototype sampling to the cost of additional training of other methods: TCL requires 2688, GroupViT 10752, and OVSegmentor 624 TFLOP/s-hours.\footnote{Estimated as training time $\times$ num. GPUs $\times$ theoretical peak TFLOP/s for GPU type.}
While training has an upfront compute cost and requires special infrastructure (\eg OVSegmentor uses 16$\times$A100s), OVDiff's prototype set can be grown progressively as needed, while showing better performance.

We additionally measure the speed of inference at 0.6s per image, which is slightly slower but comparable to 0.2s for TCL and 0.08s for OVSegmentor. We performed inference measurements using SD on the same machine with a 2080Ti GPU using 21 classes and the same resolution/sliding window settings for all methods.

\subsection{Interaction with ChatGPT}\label{sec:app_chatgpt}
\begin{table}[t]
    \centering \scriptsize 
    \caption{\textbf{Response from interaction with ChatGPT.} We used ChatGPT model to automatically categorise classes in ``stuff'' or ``things''.}
    \label{tab:chatgpt}
    \begin{tabular}{@{}l@{\hspace{2pt}}r@{\hspace{5pt}}l@{}r@{\hspace{5pt}}l@{\hspace{2pt}}r@{}}
airplane: & thing & window: & thing & awning: & thing \\
bag: & thing & wood: & stuff & streetlight: & thing \\
bed: & thing & windowpane: & thing & booth: & thing \\
bedclothes: & stuff & earth: & thing & television receiver: & thing \\
bench: & thing & painting: & thing & dirt track: & thing \\
bicycle: & thing & shelf: & thing & apparel: & thing \\
bird: & thing & house: & thing & pole: & thing \\
boat: & thing & sea: & thing & land: & thing \\
book: & thing & mirror: & thing & bannister: & thing \\
bottle: & thing & rug: & thing & escalator: & thing \\
building: & thing & field: & thing & ottoman: & thing \\
bus: & thing & armchair: & thing & buffet: & thing \\
cabinet: & thing & seat: & thing & poster: & thing \\
car: & thing & desk: & thing & stage: & thing \\
cat: & thing & wardrobe: & thing & van: & thing \\
ceiling: & stuff & lamp: & thing & ship: & thing \\
chair: & thing & bathtub: & thing & fountain: & thing \\
cloth: & stuff & railing: & thing & conveyer belt: & thing \\
computer: & thing & cushion: & thing & canopy: & thing \\
cow: & thing & base: & thing & washer: & thing \\
cup: & thing & box: & thing & plaything: & thing \\
curtain: & stuff & column: & thing & swimming pool: & thing \\
dog: & thing & signboard: & thing & stool: & thing \\
door: & thing & chest of drawers: & thing & barrel: & thing \\
fence: & stuff & counter: & thing & basket: & thing \\
floor: & stuff & sand: & thing & waterfall: & thing \\
flower: & thing & sink: & thing & tent: & thing \\
food: & thing & skyscraper: & thing & minibike: & thing \\
grass: & stuff & fireplace: & thing  & cradle: & thing \\
ground: & stuff & refrigerator: & thing & oven: & thing \\
horse: & thing & grandstand: & thing & ball: & thing \\
keyboard: & thing & path: & thing & step: & stuff \\
light: & thing & stairs: & thing & tank: & thing \\
motorbike: & thing & runway: & thing & trade name: & stuff \\
mountain: & stuff & case: & thing & microwave: & thing \\
mouse: & thing & pool table: & thing & pot: & thing \\
person: & thing & pillow: & thing & animal: & thing \\
plate: & thing & screen door: & thing & lake: & stuff \\
platform: & stuff & stairway: & thing & dishwasher: & thing \\
plant: & thing & river: & thing & screen: & thing \\
road: & stuff & bridge: & thing & blanket: & stuff \\
rock: & stuff & bookcase: & thing & sculpture: & thing \\
sheep: & thing & blind: & thing & hood: & thing \\
shelves: & thing & coffee table: & thing & sconce: & thing \\
sidewalk: & stuff & toilet: & thing & vase: & thing \\
sign: & thing & hill: & thing & traffic light: & thing \\
sky: & stuff & countertop: & thing & tray: & stuff \\
snow: & stuff & stove: & thing & ashcan: & thing \\
sofa: & thing & palm: & thing & fan: & thing \\
table: & thing & kitchen island: & thing & pier: & thing \\
track: & stuff & swivel chair: & thing & crt screen: & thing \\
train: & thing & bar: & thing & bulletin board: & thing \\
tree: & thing & arcade machine: & thing & shower: & thing \\
truck: & thing & hovel: & thing & radiator: & thing \\
monitor: & thing & towel: & thing & glass: & stuff \\
wall: & stuff & tower: & thing & clock: & thing \\
water: & stuff & chandelier: & thing & flag: & thing \\
    \end{tabular}
\end{table}
We interact with ChatGPT to categorise classes into ``stuff'' and ``things'' for the stuff filter component. Due to input limits, the categories are processed in blocks. Specifically, we input ``\textit{In semantic segmentation, there are "stuff" or "thing" classes. Please indicate whether the following class prompts should be considered "stuff" or "things":}''. We show the output in \cref{tab:chatgpt}. Note there are several errors in the response, \eg \texttt{glass}, \texttt{blanket}, and \texttt{trade name} are actually instances of tableware, bedding and signage, respectively, so should more appropriately be treated as ``things''. Similarly, \texttt{land} and \texttt{sand} might be more appropriately handled as ``stuff'', same as \texttt{snow} and \texttt{ground}. Despite this, We find ChatGPT contains sufficient knowledge when prompted with "in semantic segmentation". We have estimated the accuracy of ChatGPT in thing/stuff classification using the categories of COCO-Stuff, which are defined as 80 "things" and 91 "stuff" categories. ChatGPT achieves an accuracy rate of 88.9\% in this case.
We also measure the impact the potential errors have on our performance by providing ``oracle" answers on the Context dataset. We measure 29.6 mIoU, which is similar to $29.7\pm0.3$ of using ChatGPT, showing that small errors do not drastically affect the method, however, enable using ``stuff" filter component, which improves performance (see Table 3).

\end{document}